\documentclass{article}

\usepackage[final]{neurips_2022}

\usepackage[utf8]{inputenc} 
\usepackage[T1]{fontenc}    
\usepackage{hyperref}       
\usepackage{url}            
\usepackage{booktabs}       
\usepackage{amsfonts}       
\usepackage{nicefrac}       
\usepackage{microtype}      
\usepackage{xcolor}         

\usepackage{microtype}
\usepackage{graphicx}
\usepackage{subcaption}
\usepackage{booktabs} 

\usepackage{amsmath}
\usepackage{amssymb}
\usepackage{mathtools}
\usepackage{amsthm}
\usepackage{algorithm}
\usepackage{algorithmic}

\def\bi#1{\hbox{\boldmath{$#1$}}}

\title{Deterministic Langevin Monte Carlo with Normalizing Flows for Bayesian Inference}

\author{Richard D.P. Grumitt \\
Department of Astronomy, Tsinghua University,\\ Beijing 100084, China\\
\And
Biwei Dai\\
Physics Department, University of California\\
Berkeley,  CA 94720, USA\\
\And
Uro\v{s} Seljak \\
Physics Department, University of California\\ and Lawrence Berkeley National Laboratory\\ Berkeley,  CA 94720, USA}

\begin{document}

\maketitle

\begin{abstract}
We propose a general purpose Bayesian inference algorithm for 
expensive likelihoods, 
replacing the stochastic term in the Langevin equation with a deterministic density gradient term. 
The particle density is evaluated from the current particle positions using a Normalizing Flow (NF), which is differentiable and has good generalization properties in high dimensions.
We take advantage of NF preconditioning
and NF based Metropolis-Hastings updates for a faster convergence.  
We show on various examples that the method is competitive against state of the art sampling methods. 
\end{abstract}

\section{Introduction}
\label{Intro}

The task of Bayesian inference is to
determine the posterior $p(\bi{x}|\bi{y})$ of $d$ parameters $\bi{x}$ given the likelihood $p(\bi{y}|\bi{x})$ of some data $\bi{y}$ and 
given the prior $p(\bi{x})$, such that the posterior is given by Bayes theorem as  
$p(\bi{x}|\bi{y})= p(\bi{y}|\bi{x})p(\bi{x})/p(\bi{y})$. 
While we have access to the joint 
$p(\bi{x},\bi{y})=p(\bi{y}|\bi{x})p(\bi{x})$, the normalization 
$p(\bi{y})$ is generally 
unknown. 
Bayesian posterior inference can 
be related to the general task of finding 
a stationary distribution in an external 
potential 
by defining $U(\bi{x})=-\ln p(\bi{y}|\bi{x})-\ln p(\bi{x})=\mathcal{L}(\bi{x})+\mathcal{P}(\bi{x})$, where $\mathcal{L}(\bi{x})$ is the negative log likelihood and $\mathcal{P}(\bi{x})$ the negative log prior. 
A common 
approach
to this task is to use 
a particle
based Langevin equation, which solves
a stochastic differential 
equation (SDE) for the particle evolution
(Langevin Monte Carlo, LMC). 
 A related method is Hamiltonian Monte Carlo (HMC) \citep{duane1987hybrid}, which uses particle
positions and velocities to evolve
the particles. 
In these cases the long time equilibrium solution  is that of the 
target distribution, 
the posterior $p(\bi{x}|\bi{y})$.
LMC and HMC are two of the most popular gradient based 
implementations of 
Markov Chain Monte Carlo (MCMC), but for 
finite step size they
require Metropolis-Hastings (MH) adjustment to 
become unbiased. In this case they have
theoretical convergence 
guarantees to the stationary distribution (under mild 
ergodicity assumptions), but 
suffer from chain element 
correlations, which may render 
the convergence to be very slow.


In this paper we propose a Deterministic
Langevin Monte Carlo (DLMC) with Normalizing Flows (NF). It uses a  
random initialization
from the prior, but the subsequent evolution is deterministic, 
replacing the stochastic velocity 
term in the Langevin equation with 
a deterministic density gradient 
term. 
To avoid the need to solve the Fokker-Planck equation directly 
we evaluate  
the density defined by the current particle
positions via
an NF, and then use its gradient to 
update the first order particle Langevin
dynamics. NFs are differentiable and have good generalization properties in high dimensions.

Deterministic sampling methods have 
until now been limited to 
particle based methods. Two examples are 
Stein Variational Gradient Descent (SVGD) \citep{Liu2016stein}, and  
interacting particle 
solutions of
\cite{2020Entrp..22..802M}, which proposes 
particle based scores for the gradient of the log-density. However, 
these particle based methods scale poorly to 
high dimensions.  The main 
novelty of our contribution 
is the introduction of 
NFs to the class of deterministic methods for evaluating
stationary distributions: we use NFs for density 
estimation and gradient evaluation, as 
well as for MH adjustment to correct for imperfect NF density estimation and accelerate convergence. 


Our primary target application is to expensive likelihoods (wall clock time of seconds or more). An example are inverse problems in scientific  
applications, where often one must solve an expensive Ordinary or Partial Differential Equation (ODE or PDE)
forward model, where the computational cost 
can be minutes or hours. 
In such applications the computational 
cost of the NF (which is of order seconds in our experiments) is not dominant. For fast model evaluations standard MCMC methods
suffice and our target are problems where 
the cost of standard methods is prohibitive. 
The computational cost of a forward model is unrelated to the 
complexity of the posterior distribution: often 
the posteriors are simple Gaussian distributions
even when the computational cost is high. In such settings the main goal of posterior 
analysis is to minimize the number of 
likelihood calls to reach some prescribed
precision on the posterior. 

DLMC has the following advantages in comparison to several other MCMC samplers:
\begin{itemize}
\item DLMC takes advantage of gradient 
information of the target distribution, but without stochastic 
noise, which slows 
down the particle evolution towards the 
target in LMC and HMC. As a result, DLMC gradient based optimization can rapidly move the particles into 
the region of high posterior mass for a fast convergence. 
\item DLMC is not a sequential Markov Chain, and DLMC particles can be updated in 
parallel at each time step, which can take advantage of trivial machine parallelization of likelihood evaluations, reducing 
the wall-clock time compared to standard MCMC. 
\item
DLMC particles are initialized as a random draw from the prior, which covers the posterior. Each particle produces an independent sample in the limit of a large number of particles ($N$). 
\item
DLMC uses NF for density estimation, and takes advantage of NF preconditioning and NF based MH updates, for a faster convergence and to help correct for imperfect density estimation. 
\item
DLMC can handle multimodal posteriors, in contrast to standard MCMC samplers.
\end{itemize}

\section{Langevin and Fokker-Planck equations}\label{sec: langevin_fp}

The overdamped Langevin 
equation is a stochastic differential 
equation describing particle 
motion in an external potential and 
subject to a random force with zero mean,
\begin{equation}
  \dot{\bi{x}}=\bi{v}=-\bi{\nabla} U(\bi{x})+\bi{\eta},
    \label{Lan3}
\end{equation}
where $\langle \bi{\eta(t)}\rangle=0$ and $\langle \eta_i(t) \eta_j(t') \rangle=2\delta_{ij}\delta(t-t')$. 
For simplicity we set the diffusion coefficient, 
temperature and mobility to unity.
The Langevin 
equation is a first order velocity equation which has a deterministic velocity $-\bi{\nabla} U(\bi{x})$ 
and a stochastic 
velocity $\bi{\eta}$. Here the gradient operator is with respect to the 
parameters of $U(\bi{x})$, $\bi{\nabla}U(\bi{x})=dU(\bi{x})/d\bi{x}$.

In practice we need to discretize the Langevin 
equation, and 
for any finite step size $\Delta t$ the result is 
a biased distribution. 
An example is the Ornstein-Uhlenbeck 
process, where $U(\bi{x})$ is a
harmonic potential, $\exp(-U(\bi{x}))=N(\bi{x};\bi{\mu},\bi{\Sigma})
$. Standard Langevin algorithm updates lead to 
the solution $q(\bi{x})=N(\bi{\mu},\bi{\Sigma}(I-\Delta t \bi{\Sigma}^{-1}/2)^{-1})$ \citep{wibisono2018sampling}. 
We can see that for a finite fixed 
stepsize $\Delta t$ the solution 
converges to a biased answer regardless of the number 
of time integration steps taken, or the number of samples used. 
A solution to this is to 
supplement the Langevin updates with the
Metropolis-Hastings (MH) Adjustment Langevin Algorithm (MALA), where 
MH acceptance guarantees detailed 
balance \citep{roberts1996exponential}. 
A similar MH adjustment is also required for 
HMC \citep{duane1987hybrid}.

The Langevin equation can be viewed as a particle 
implementation of the evolution of the
particle probability density $q(\bi{x}(t))$, which is governed by the deterministic
Fokker-Planck equation, a continuity equation for the density,
\begin{equation}
    \dot{q}(\bi{x}(t))+\bi{\nabla}\cdot \bi{J}=0,\quad \bi{J}=-q(\bi{x}(t))[\bi{\nabla} U(\bi{x}(t))-\bi{\nabla}  V(\bi{x}(t))]\equiv q(\bi{x}(t))\bi{v}. 
    \label{fp}
\end{equation}
We 
defined $V(\bi{x}(t))=-\ln q(\bi{x}(t))$ and 
expressed current as density times velocity, 
where the two terms in the probability current $\bi{J}$ correspond
to the two velocity terms in the Langevin equation.
One can see from equation 
\ref{fp} that the stationary 
distribution $\dot{q}(\bi{x},t)=0$ 
is given by  $ p(\bi{x}|\bi{y})=\exp(-U(\bi{x}))/p(\bi{y})$.
The Fokker-Planck equation is a PDE, 
while the Langevin equation is stochastic (SDE). 
Langevin diffusion provides a particle 
description of the density ${q}(\bi{x},t)$ and in the 
large time limit both 
equations lead to a stationary 
distribution equal to the posterior $p(\bi{x}|\bi{y})$. 



If we replace the
stochastic velocity in 
the Langevin equation \ref{Lan3} 
with the deterministic velocity in 
equation \ref{fp}, we obtain the
deterministic Langevin equation \citep{2020Entrp..22..802M,song2020score},
\begin{equation}
    \dot{\bi{x}}(t)=\bi{v}=-\bi{\nabla} [U(\bi{x}(t))-V(\bi{x}(t))]. 
    \label{Lan4}
\end{equation}
Equation \ref{Lan4} gives the dynamics
of a particle representation of the density $q(\bi{x}(t))$, which in the large time limit $t\rightarrow \infty$ converges to the same
distribution as the solution of equation \ref{Lan3}. This follows from the 
fact that the stationary solution of equation \ref{Lan4},
$\dot{\bi{x}}(t)=0$ is given by $V(\bi{x}(t))= U(\bi{x}(t))+\rm{c}$, 
with c being a constant independent 
of $\bi{x}$, which must thus be equal 
to $\log p(\bi{y})$, since $q(\bi{x}(t))$ is normalized. There is no 
stochastic noise in Equation \ref{Lan4}, so it can be interpreted
as a deterministic analog of the Langevin equation, which is a first order equation for velocity. 

The main difficulty in solving equation \ref{Lan4} is the instantaneous density 
term $q(\bi{x}(t))$ needed to obtain $V(\bi{x}(t))$. Here we solve this 
by evolving $N$ particles together, 
using NFs to evaluate their particle density. 
While this makes the particles interacting, 
in the limit of large $N$ the particle 
evolution is independent. 
Furthermore, we can use the NF to do MH adjustment 
to correct for imperfect NF density estimation (Section \ref{sec3}). 
Upon time discretization of Equation \ref{Lan4} 
we obtain  
Algorithm \ref{alg1}. A detailed discussion of the algorithm implementation is given in Appendix \ref{app: dlmc_algo}.  
We initialize at $t=0$ by providing initial 
samples drawn at random
from the known prior distribution $p(\bi{x})$, similar to Sequential Monte Carlo \citep{del2006sequential}.
This  has several positive features: 
since the posterior is proportional to the
prior times the likelihood, the prior always covers the
posterior. Moreover, when the likelihood is weakly 
informative the convergence of DLMC is fast. 
We then apply DLMC gradient updates to 
obtain the particle position updates. When 
initializing from the prior the 
initial density distribution cancels
out the prior in the target, and 
the initial gradient update is 
the log likelihood term $\bi{\nabla} \mathcal{L}(\bi{x})$.

Equation \ref{Lan4} can be interpreted as a 
gradient based minimization of the time 
dependent objective $U(\bi{x}(t))-V(\bi{x}(t))$. 
In this view we can optimize the objective  using any optimization method. 
Initially, if we 
start the particles from the prior $p(\bi{x})$, DLMC is simply 
performing
optimization of the target likelihood $p(\bi{y}|\bi{x})$. Later, as 
$q(\bi{x}(t)) \sim p(\bi{x}|\bi{y})$, the gradient of the optimization 
objective becomes zero, and the particles
stop moving. One can use any optimization method to 
move the particles, so if the gradient of $U(\bi{x})$
is not available we can use gradient free optimization methods to do so. For gradient 
based optimization we can use many of 
the standard optimization methods \citep{NocedalWright06}, such as first order (stochastic or deterministic) gradient descent, 
momentum based methods, or  (quasi) second order methods (e.g., L-BFGS, Newton, conjugate gradient etc.). 
The algorithm can be stopped when the 
particle distribution becomes stationary. 
For example, 
we can stop when the estimated first and second 
moments of the posterior stop changing. 

Given the formulation above, the performance of DLMC depends on the ability of the NF to approximate the particle density and the corresponding gradient at each time step. The addition of MH updates (Section \ref{sec3}) help to correct for imperfect density estimation. However, in situations where the NF fit is a poor match for the target the MH acceptance can be very low. Furthermore, in high dimensions ($\mathcal{O}(10^3)$) we are limited by the training cost to use simple NF approximations. However, in this scenario DLMC can still be used to move particles quickly towards the typical set, acting as an initialization for MCMC in the NF latent space. This is discussed further in Section \ref{sec: ill_gauss}. 


\section{Normalizing Flow density estimation}
Normalizing Flows provide a powerful framework for density estimation and sampling \citep{dinh2016density, papamakarios2017masked,kingma2018glow,dai2021sliced}. These models map the $d$-dimensional data $\bi{x}$ to $d$-dimensional latent variables $\bi{z}$ through a sequence of invertible transformations $f = f_1 \circ f_2 \circ ... \circ f_L$, such that $\bi{z} = f(\bi{x})$ and $\bi{z}$ is mapped to a base distribution $\pi(\bi{z})$, which is chosen to be a standard Normal distribution $N(0,\bi{I})$. The probability density of data $\bi{x}$ can be evaluated using the change of variables formula:
\begin{align}
    \label{eq:flow}
    q(\bi{x})\equiv  e^{-V(\bi{x})} = \pi(f(\bi{x})) \left|\det \left(\frac{\partial f(\bi{x})}{\partial \bi{x}}\right)\right| 
    = \pi(f(\bi{x})) \prod_{l=1}^L \left|\det \left(\frac{\partial f_l(\bi{x})}{\partial \bi{x}}\right)\right| .
\end{align}
The Jacobian determinant of each transform $J_l= |\det (\frac{\partial f_l(\bi{x})}{\partial \bi{x}})|$ must be easy to compute for evaluating the density, and the transformation $f_l$ should be easy to invert for efficient sampling.
In this paper we use Sliced Iterative Normalizing Flow (SINF) \citep{dai2021sliced}, which 
has been shown to achieve better performance for 
small training data (below a few thousand particles), and is considerably faster, than 
alternatives. 

\subsection{Normalizing Flow as a  preconditioner}

While the core of DLMC is the NF based density gradient, the 
same NF can be used for additional 
convergence acceleration towards the target. 
The proposed algorithm can be slow 
when the condition number of the problem is high. 
A typical solution is to precondition with a covariance matrix, which mimics the global curvature of the problem. This becomes more difficult if the curvature is position dependent. 
NFs can  address 
this problem by warping the space 
in which we perform the optimization. This warping maps spatially varying 
curvature to the latent space where the curvature is 
constant and isotropic \citep{parno2018transport,hoffman2019neutra}. 
We can implement DLMC in the 
NF latent space $\bi{z}$, where the target
distribution $p(\bi{x}|\bi{y})$ is transformed 
into
\begin{equation}
    p(\bi{z}|\bi{y})= p(\bi{f}^{-1}(\bi{z})|\bi{y})J(\bi{z}),
    \label{pz}
\end{equation}
where $J(\bi{z})=\left|\det \left(\frac{\partial \bi{f}^{-1}(\bi{z})}{\partial \bi{z}}\right)\right|$ is the absolute value of the Jacobian determinant. 
We can define the potential in 
latent space as 
\begin{equation}
{U}(\bi{z})=U(\bi{f}^{-1}(\bi{z}))-\ln J(\bi{z}).
\end{equation}
If we can evaluate $\bi{\nabla{U}}(\bi{x})$ using methods such as auto-differentiation 
and if the NF Jacobian is also differentiable we can also 
evaluate the gradient in latent space $\bi{z}$ as 
\begin{equation}
\bi{\nabla}U(\bi{z})= \bi{\nabla} U(\bi{x}) \frac{d\bi{x}}{d\bi{z}}-\bi{\nabla} \ln J(\bi{z})
\end{equation}

In latent space the latent variables are 
distributed as $\pi(\bi{z})=\exp(-V(\bi{z}))=N(0,\bi{I})$, 
so $V(\bi{z})=\bi{z}^T \bi{z}/2+(d/2)\ln(2\pi)$ and $\bi{\nabla} V(\bi{z})=\bi{z}$. This gives the deterministic Langevin
gradient descent updates as 
\begin{align}
    \bi{z}(t+\Delta t)=&\bi{z}(t)-\bi{\nabla} U(\bi{z}(t))\Delta t+\bi{z}(t)\Delta t. 
    \label{Lan2}
\end{align}
It is interesting 
to compare this deterministic update to the standard stochastic Langevin equation  update, which  
in latent space is 
\begin{align}
    \bi{z}(t+\Delta t)=&\bi{z}(t)-\bi{\nabla} U(\bi{z}(t))\Delta t+\bi{\eta}', 
    \label{Lan6}
\end{align}
where $\bi{\eta}'\sim N(0,2\bi{I}\Delta t)$. Both updates have 
the potential gradient term, which is the deterministic drift velocity. 
The difference between the two is  that DLMC also
uses the current position of the sample $\bi{z}$ 
for a deterministic update of the density 
gradient $\bi{z}(t)\Delta t$, which 
moves the particles radially outwards from 
the center of the latent space, 
while the stochastic Langevin algorithm adds a Gaussian random 
variable $\bi{\eta}'$ to the update. We find that performing DLMC updates in the NF latent space accelerates convergence. 


\begin{algorithm}[tb]
   \caption{Deterministic Langevin Monte Carlo with Normalizing Flows}
\begin{algorithmic}[1]
   \STATE {\bfseries Input:} initial samples $\bi{x}^{i}(t=0)$ drawn at random from prior $p(\bi{x})$, size $N$,  access to $\mathcal{L}(\bi{x})+\mathcal{P}(\bi{x})=U(\bi{x})$ and its gradient.
   \STATE Initial update:
   \FOR{$i=1$ to $N$}
   \STATE $\bi{x}^i(\Delta t)=\bi{x}^i(t=0)-\bi{\nabla} \mathcal{L}(\bi{x}^i(t))\Delta t$
   \ENDFOR
    \WHILE{not converged}
    \STATE $t\leftarrow t+\Delta t$
    \STATE Run Normalizing Flow on current particle positions to obtain $q(\bi{x}(t))=\exp(-V(\bi{x}(t)))$ and map $\bi{z}(t)=f(\bi{x}(t))$
     \FOR{$i=1$ to $N$}
     \IF{latent space update}
     \STATE Map to latent space: $\bi{z}^{i}(t) = f(\bi{x}^{i}(t))$
     \STATE DL update: $\bi{z}^i(t+\Delta t)=\bi{z}^i(t)-[\bi{\nabla} U(\bi{z}^i(t))-\bi{z}^i(t)]\Delta t$
     \STATE Inverse map: $\bi{x}^{i}(t+\Delta t)=f^{-1}(\bi{z}^{i}(t+\Delta t))$
     \ELSE
     \STATE DL update: $\bi{x}^i(t+\Delta t)=\bi{x}^i(t)-\bi{\nabla} [U(\bi{x}^i(t))-V(\bi{x}^i(t))]\Delta t$
     \ENDIF
     \STATE MH update: draw a sample $\tilde{\bi{x}}^i\sim q(\bi{x}(t))$, replace $\bi{x}^i$ with $\tilde{\bi{x}}^i$ with probability
    $r(\bi{x}^i,\tilde{\bi{x}}^i)$
   \ENDFOR
   \ENDWHILE
   \STATE {\bfseries Output}: a set of particles $\bi{x}^i(t)$ distributed as a target distribution $p(\bi{x}|\bi{y})$. 
\end{algorithmic}
\label{alg1}
\end{algorithm}

\subsection{Normalizing Flow for Metropolis-Hastings adjustment} 
\label{sec3}

The main requirement of DLMC is that $q(\bi{x}(t))$ 
properly describes the density of samples $\bi{x}(t)$
at time $t$. This can only be ensured
in the large $N$ limit. 
Moreover, for non-convex optimization 
problems it is possible that the samples get stuck 
at local minima  far from the 
global minimum. 
One can add a Metropolis-Hastings (MH) adjustment to correct for imperfect density estimation, eliminating samples that get stuck far from the global minimum. 
This step is particularly necessary for disconnected multi-modal peaks of the target to be 
properly equilibrated. We note that whilst LMC (i.e., MALA) and HMC utilize local MH adjustment, they typically cannot equilibriate between separate posterior peaks without the use of methods such as annealing.  

Specifically, at a given $t$ we can draw independent particles $\tilde{\bi{x}}^i \sim q(\bi{x}(t))$, one for each 
existing particle $\bi{x}^i$. We compare the new particle 
against an existing particle to decide whether we 
accept or reject it. 
Since these new samples are 
independent the 
MH acceptance rate is (e.g., \cite{albergo2019flow})
\begin{equation}
    r(\bi{x},\tilde{\bi{x}})=\min\left\{1,\frac{p(\tilde{\bi{x}},\bi{y}) q(\bi{x})}{p(\bi{x},\bi{y}) q(\tilde{\bi{x}})}\right\}.
    \label{MH}
\end{equation}
Note that the normalization constant cancels out 
in this ratio, so we can evaluate this with knowledge 
of $U(\bi{x})$ only. 

Initially, when $q(\bi{x}(t))$ is very broad compared to 
$p(\bi{x}|\bi{y})$, the acceptance rate of MH will be
low, since most of the samples drawn will 
have low $p(\bi{x}|\bi{y})$, so this process may only  eliminate the worst
performers, and replace them with samples in the 
higher density region of $p(\bi{x}|\bi{y})$. As DLMC
progresses $q(\bi{x}(t))$ becomes closer to $p(\bi{x}|\bi{y})$, 
and the acceptance increases and 
reaches $r(\bi{x},\tilde{\bi{x}})=1$
if the NF learns the target perfectly.
In high dimensions the MH acceptance can remain low. However, the MH update is  still necessary for
proper equilibration of multi-modal peaks. 


\begin{figure}[h]
     \centering
      \begin{subfigure}[h]{0.3\linewidth}
         \includegraphics[width=\textwidth]{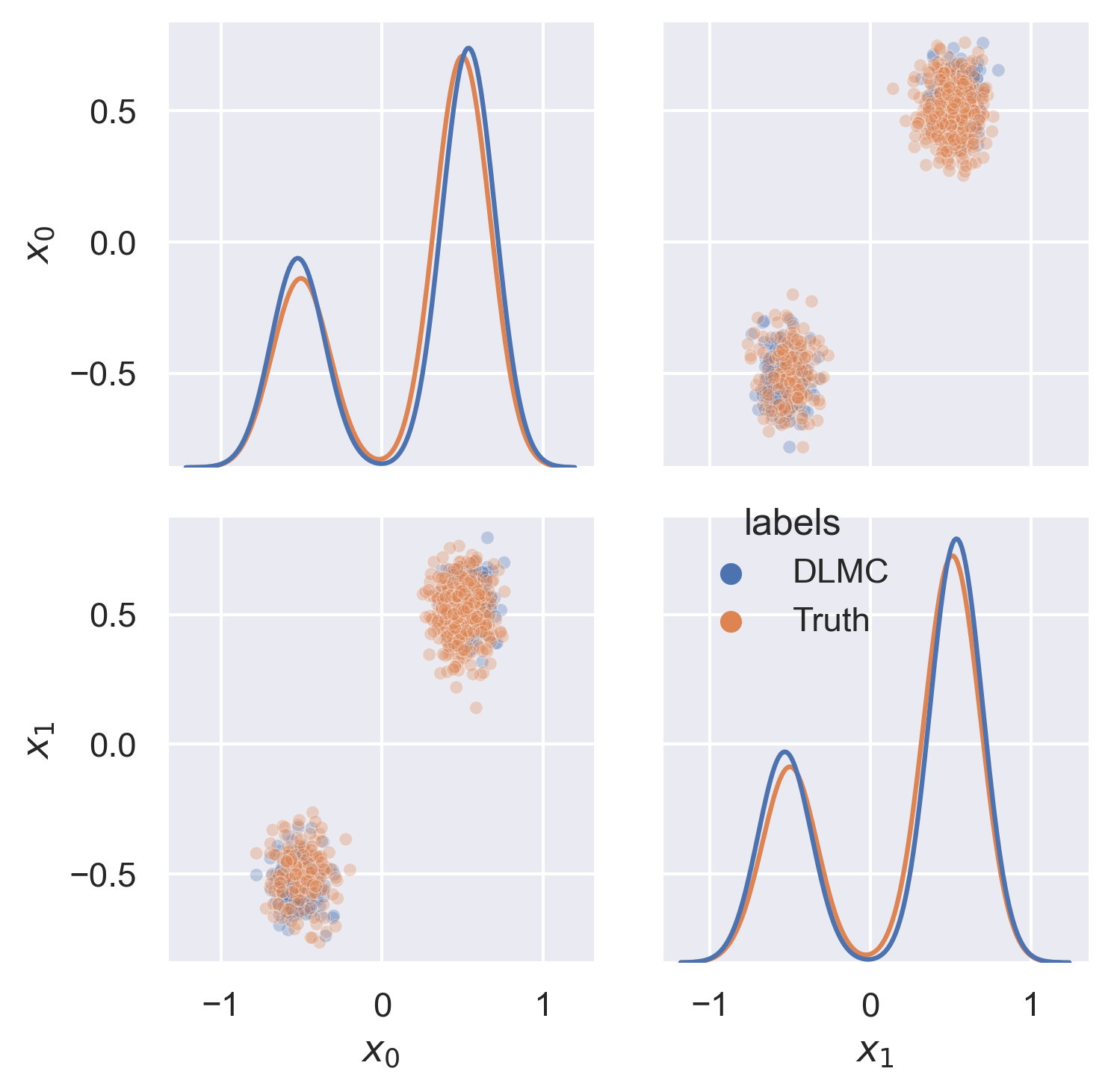}
         \caption{Double Gaussian}
     \end{subfigure}
     \begin{subfigure}[h]{0.3\linewidth}
         \includegraphics[width=\textwidth]{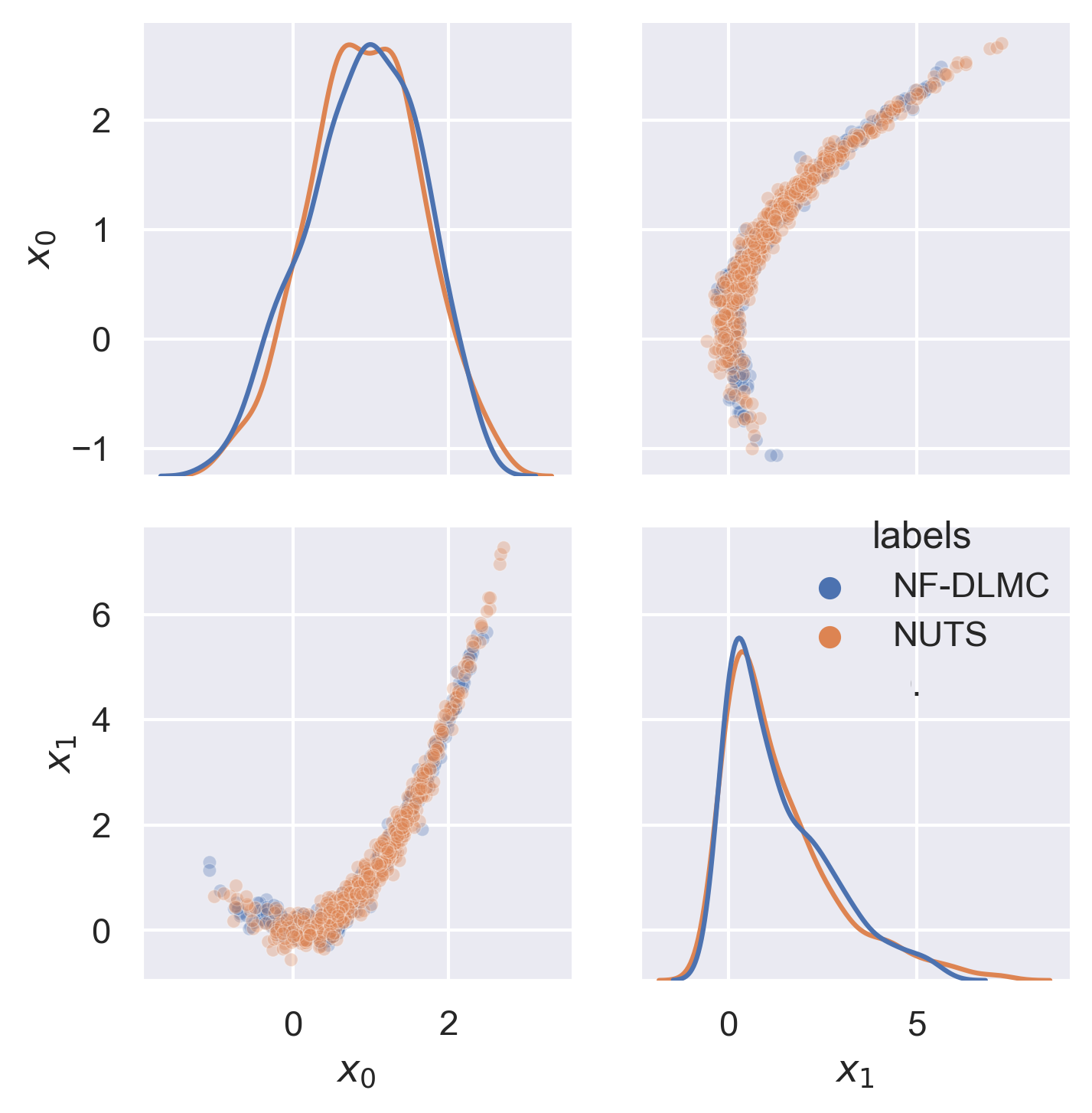}
         \caption{Rosenbrock}
     \end{subfigure}
     \begin{subfigure}[h]{0.3\linewidth}
         \includegraphics[width=\textwidth]{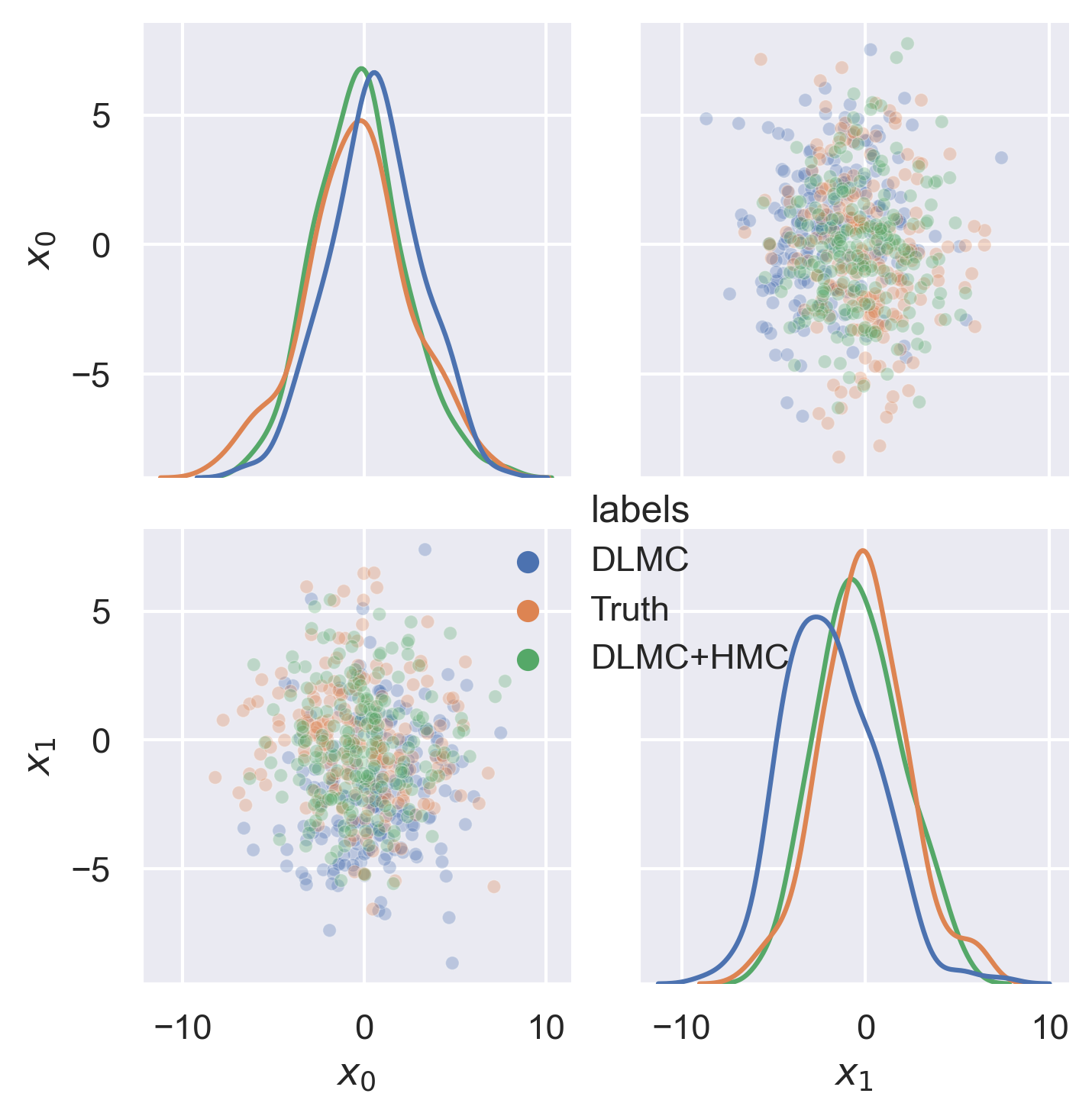}
         \caption{Ill conditioned Gaussian}
     \end{subfigure}
     \caption{Panel (a): results on $d=100$ Gaussian mixture after 90 DLMC iterations (blue), compared to the true distribution (orange). We show 1d density projections and 2d particle distributions along the first two parameters. Panel (b): same for Rosenbrock function in $d=32$ after 50 DLMC iterations. Panel (c): particle distributions for the $d=1000$ ill conditioned Gaussian after 28 DLMC iterations, along with the distribution following 5 subsequent latent space HMC iterations (green). The Gaussian mixture uses 500 particles, the Rosenbrock uses 1000, and the ill conditioned Gaussian uses 2000.}
     \label{fig2}
      \vskip -0.1in
\end{figure}

\begin{figure*}[t]
     \centering
      \begin{subfigure}[t]{0.3\linewidth}
         \includegraphics[width=\textwidth]{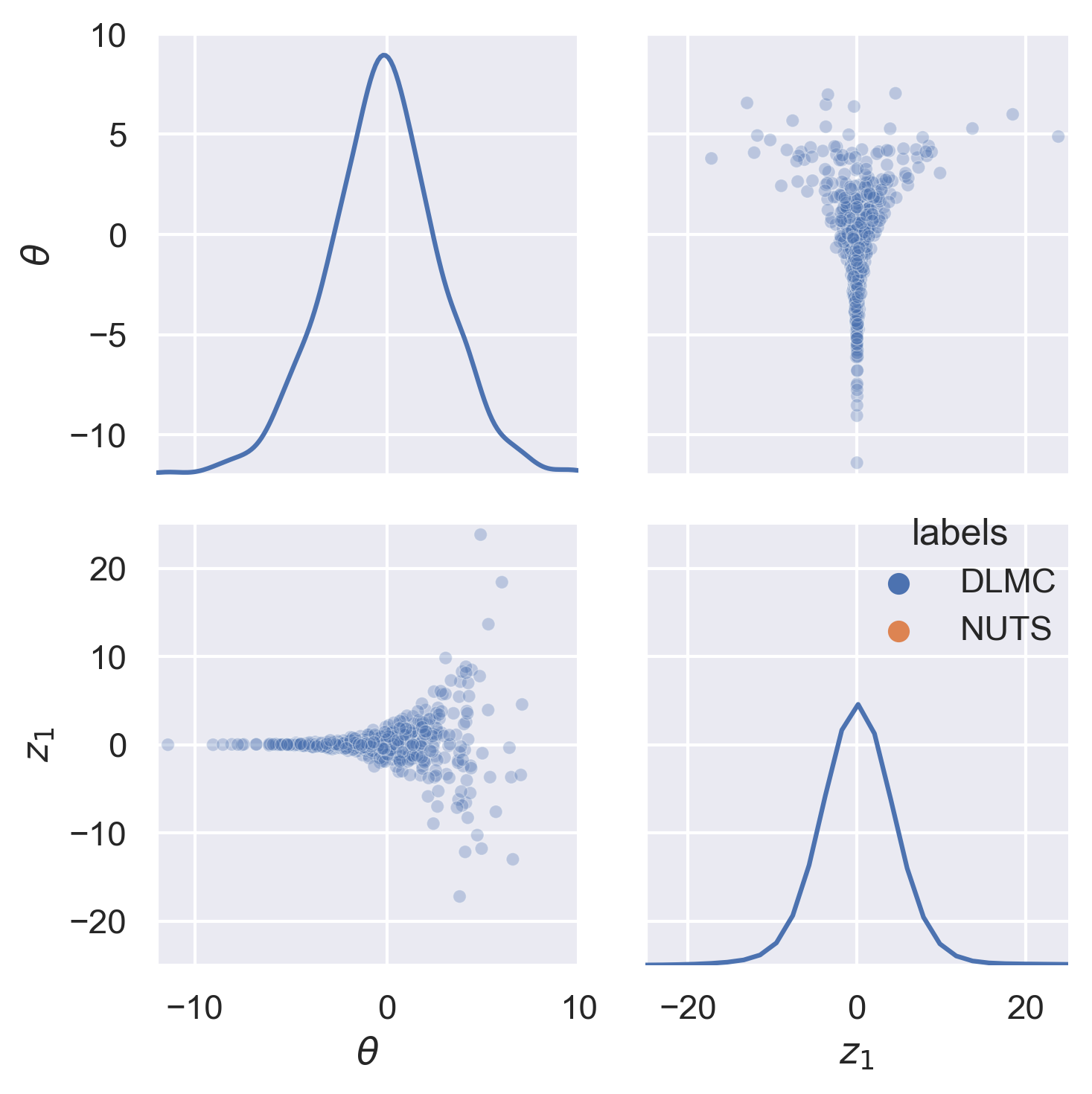}
         \caption{Neal's funnel ($\sigma=\infty$)}
     \end{subfigure}
       \begin{subfigure}[t]{0.3\linewidth}
         \includegraphics[width=\textwidth]{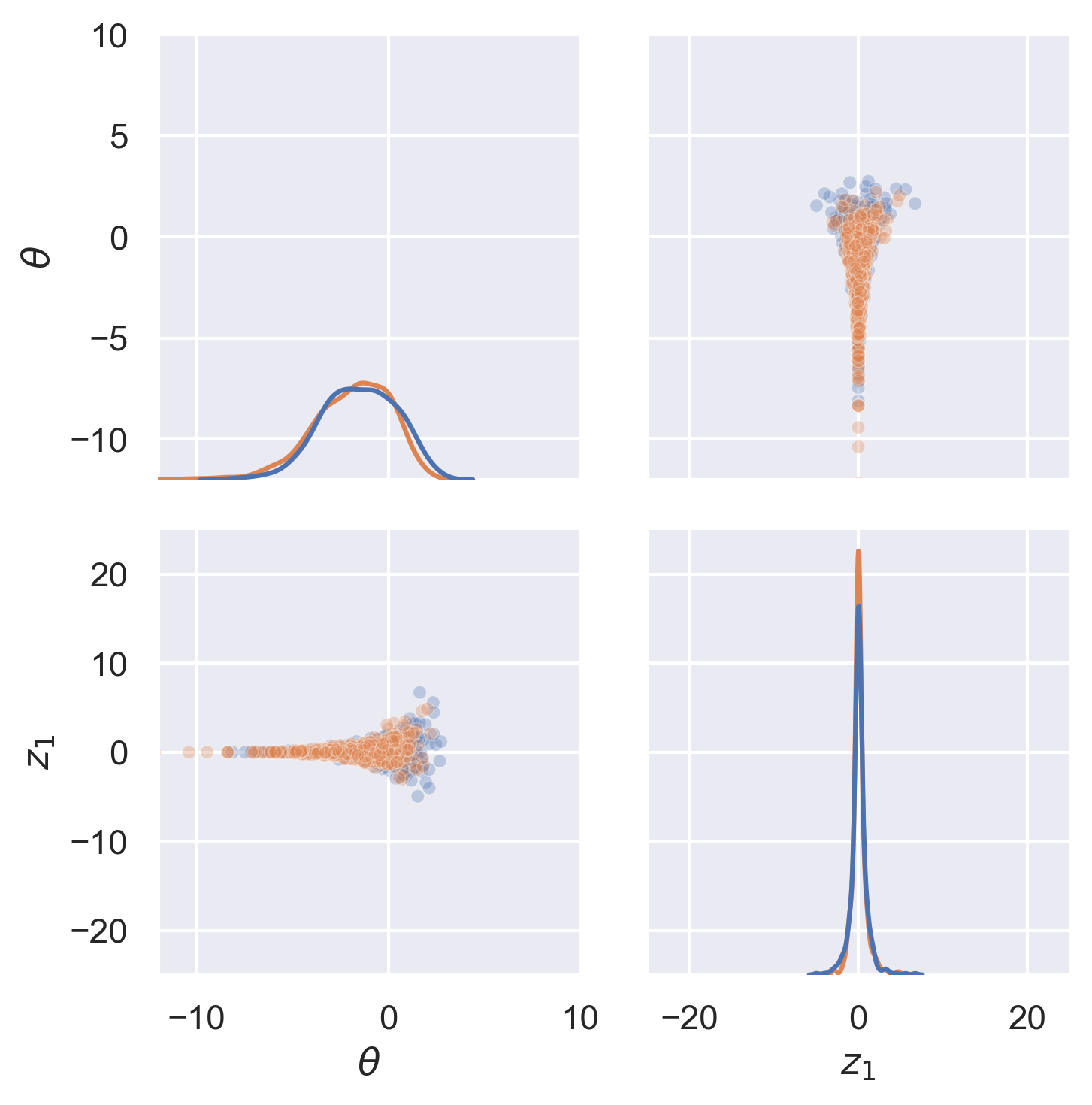}
         \caption{$\sigma=5$}
     \end{subfigure}
       \begin{subfigure}[t]{0.3\linewidth}
         \includegraphics[width=\textwidth]{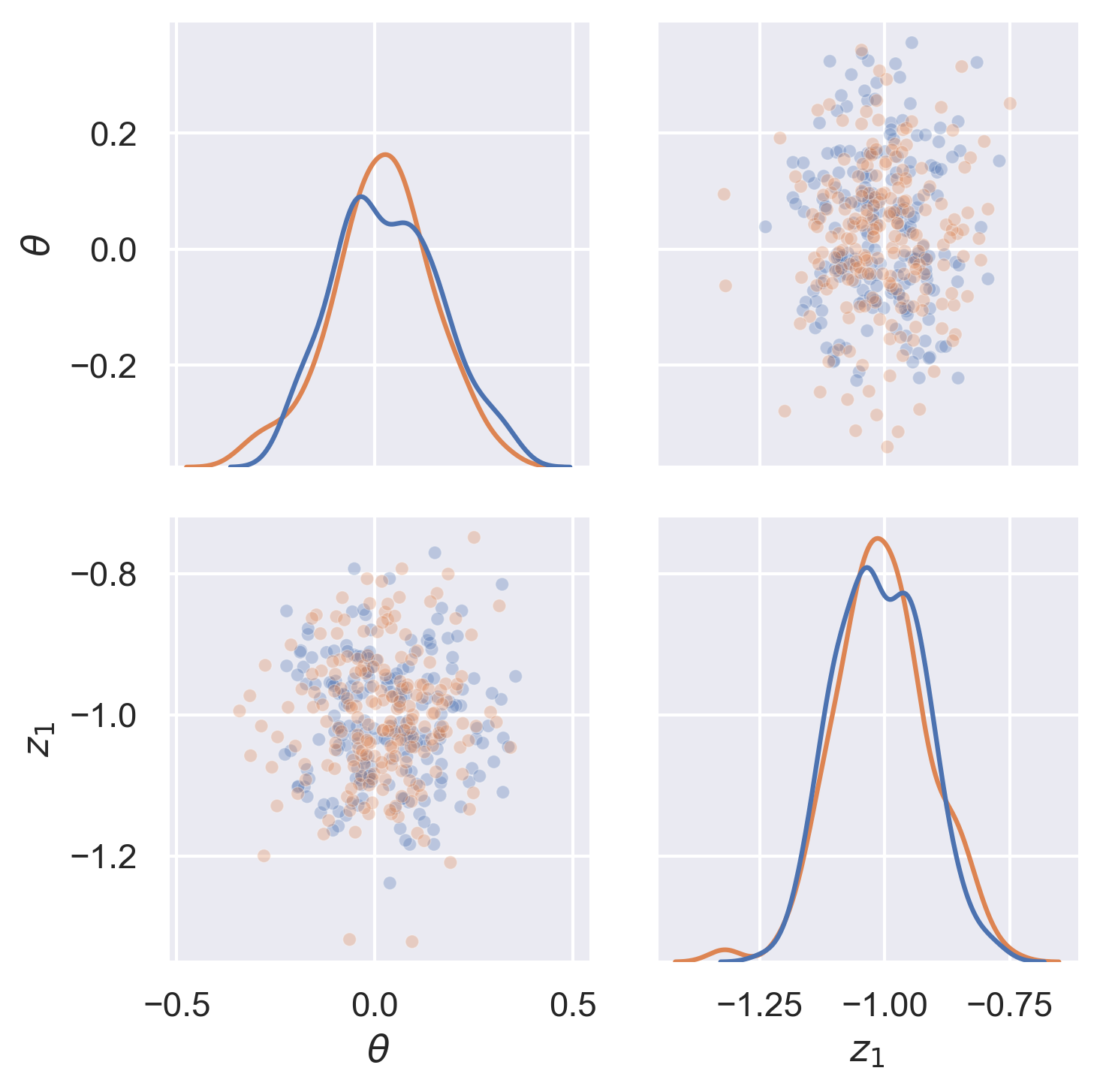}
         \caption{$\sigma=0.1$}
     \end{subfigure}
     \caption{DLMC results on $d=101$ hierarchical variance problem (funnel problem), compared to a very long NUTS run. We show posteriors of $\theta$ and $z_1$. Left panel shows Neal's funnel: it corresponds to noise $\sigma\rightarrow \infty$, so no updates are needed since the likelihood update is zero. Middle panel shows $\sigma=5$: despite the large noise, the 
     $d=100$ latent variables are informative of $\sigma$ and the posterior 
     is narrower than the prior. 
     Right panel shows $\sigma=0.1$: here the posterior is narrowed further, and is close to a Gaussian.
     }
     \label{fig3}
      \vskip -0.1in
\end{figure*}

\begin{figure*}[t]
     \centering
     \begin{subfigure}[t]{0.45\linewidth}
         \includegraphics[width=\textwidth]{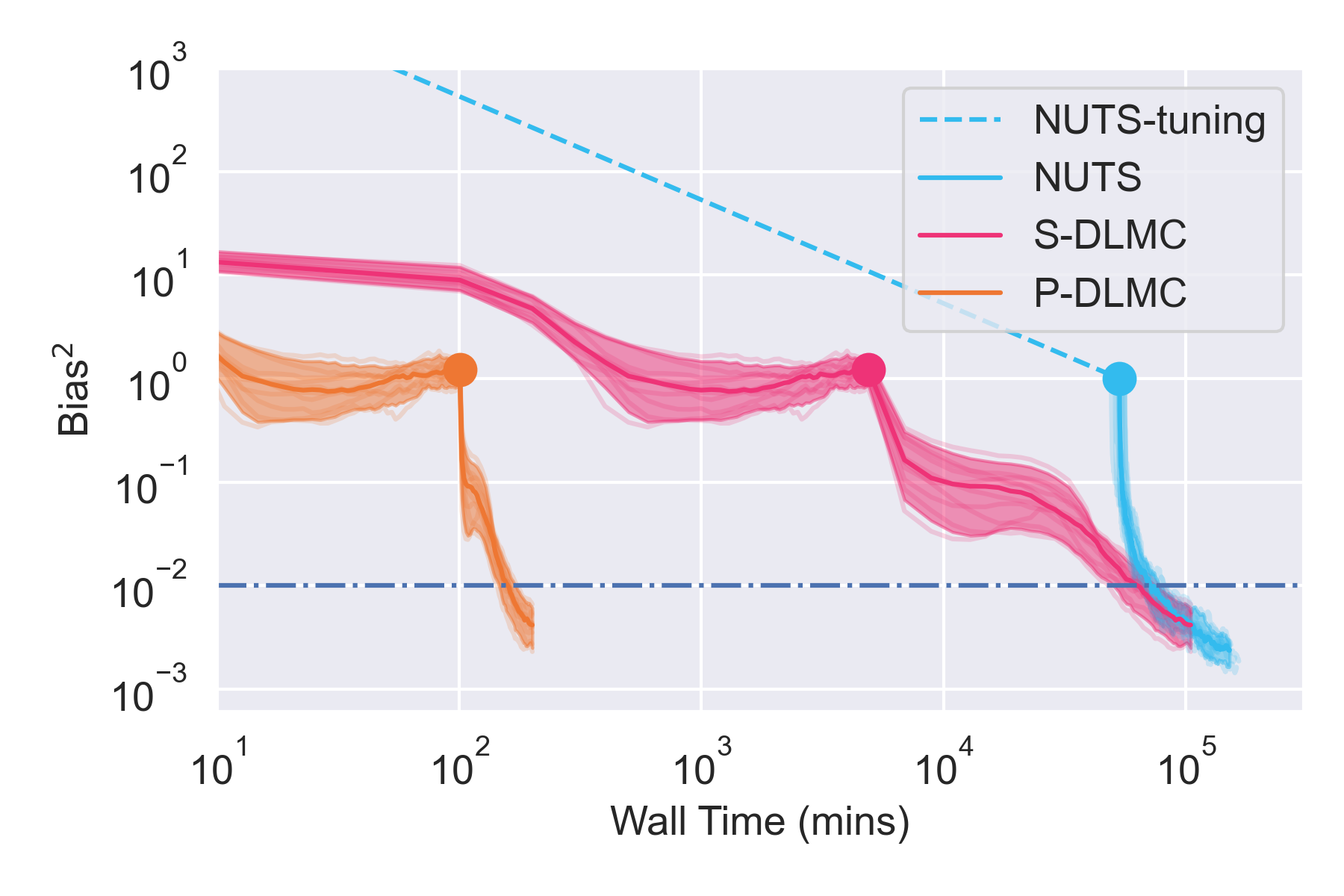}
         \caption{Rosenbrock}
     \end{subfigure}
      \begin{subfigure}[t]{0.45\linewidth}
         \includegraphics[width=\textwidth]{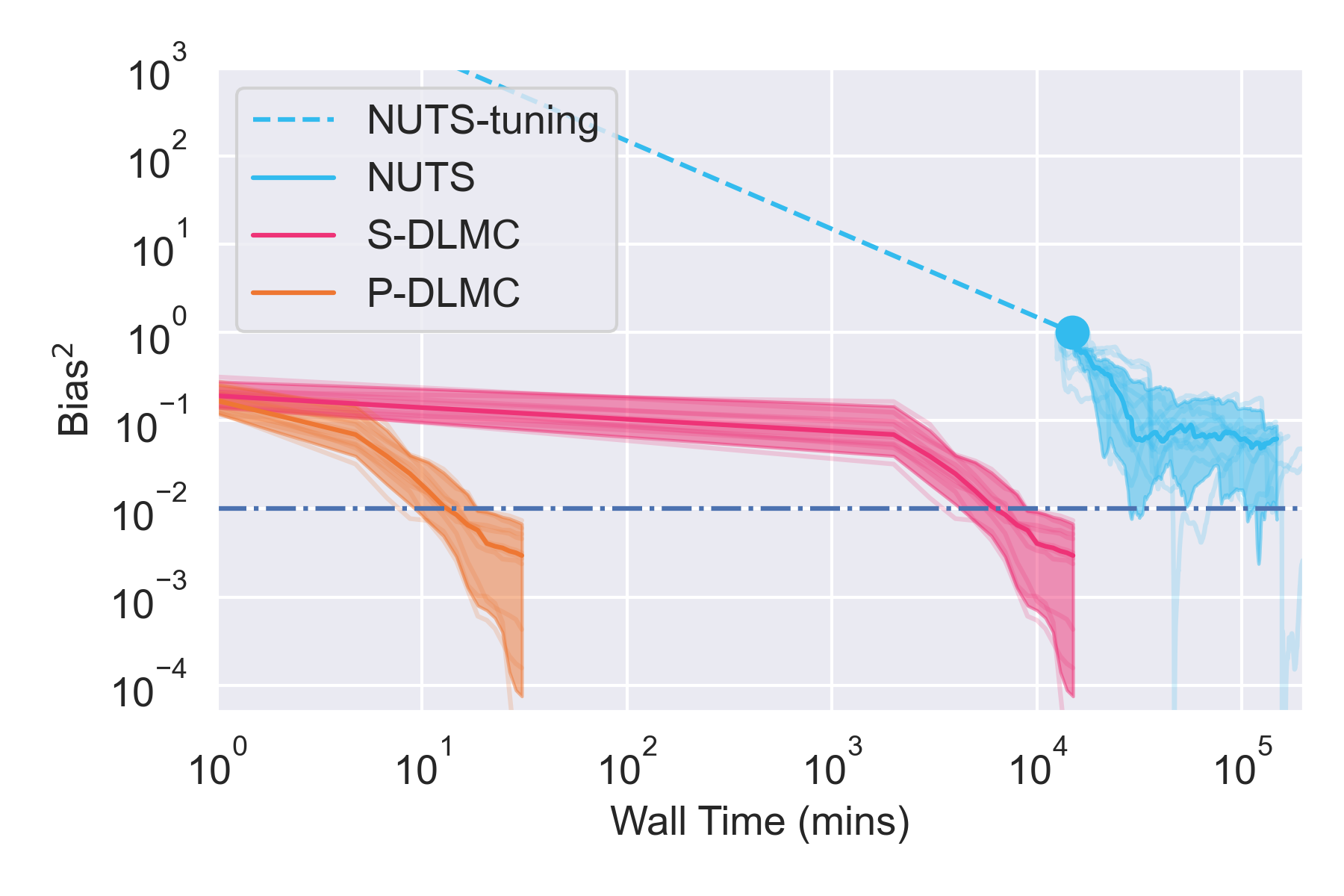}
         \caption{funnel $\sigma=5$}
     \end{subfigure}
       \begin{subfigure}[t]{0.45\linewidth}
         \includegraphics[width=\textwidth]{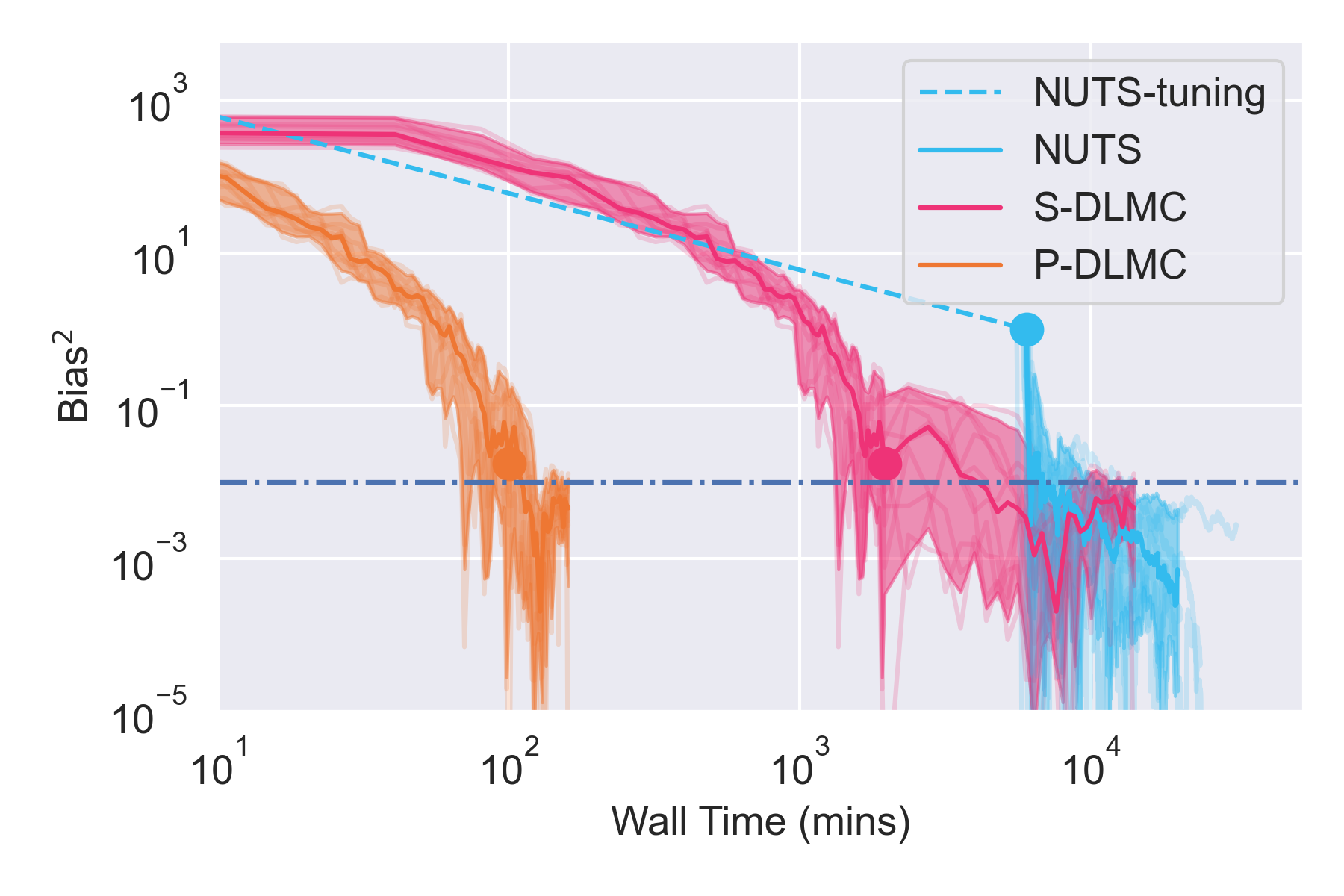}
         \caption{funnel $\sigma=0.1$}
     \end{subfigure}
      \begin{subfigure}[t]{0.45\linewidth}
         \includegraphics[width=\textwidth]{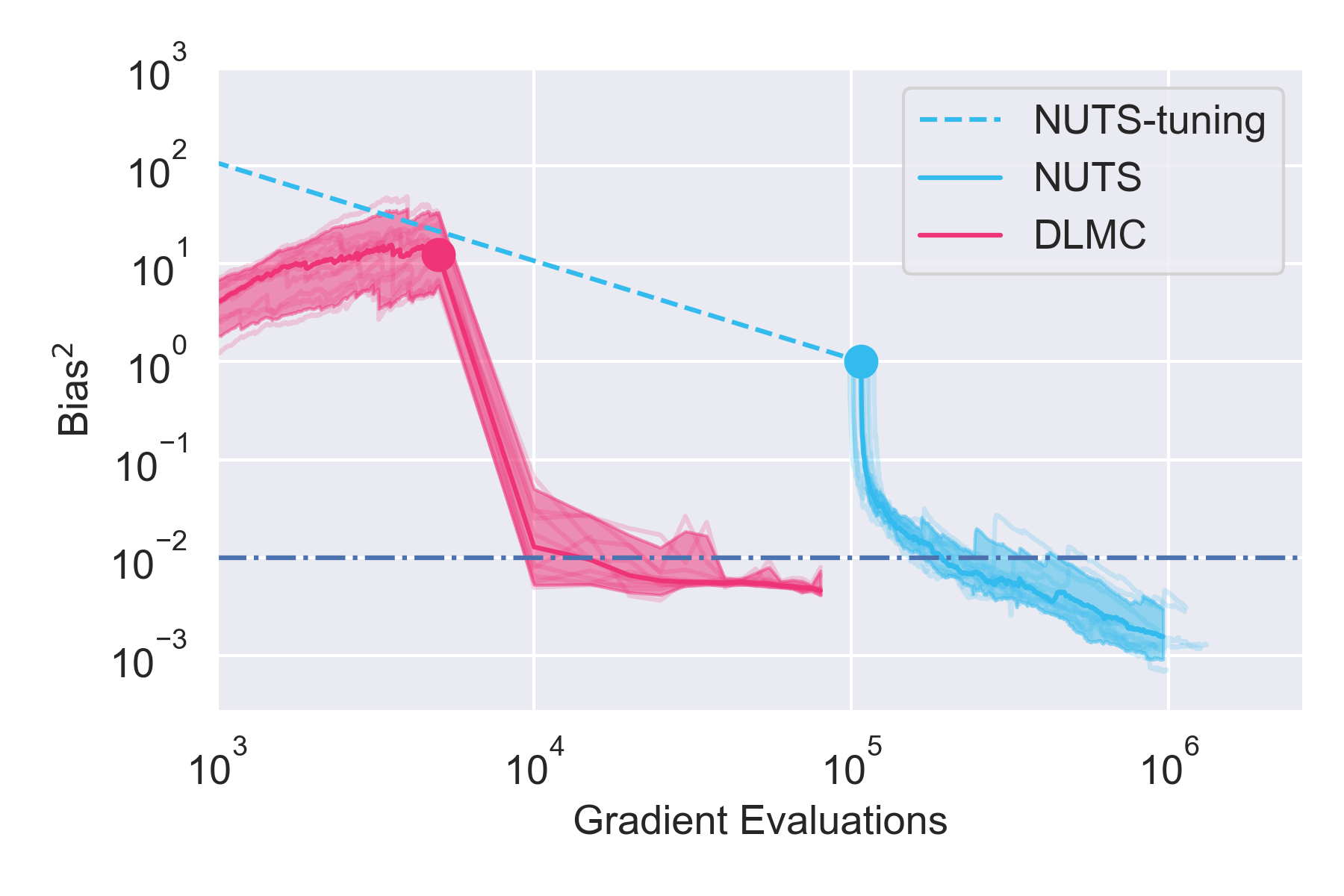}
         \caption{sparse logistic regression}
     \end{subfigure}
     \caption{Bias-squared on the second moment for the $d=32$ Rosenbrock function (panel (a)), $d=101$ hierarchical funnel vs wall clock time (panels (b) and (c)) and for $d=51$ sparse logistic regression versus number of likelihood gradient evaluations (panel (d)). In panels (a)-(c) we show serial (S-DLMC) and parallel (P-DLMC) versions. Solid circles denote the end of any burn-in phases.
     }
     \label{fig:funnel_bias2}
      \vskip -0.1in
\end{figure*}

\begin{figure}[t]
     \centering
     \hskip 0.3in
      \begin{subfigure}[t]{\linewidth}
         \includegraphics[width=0.9\textwidth]{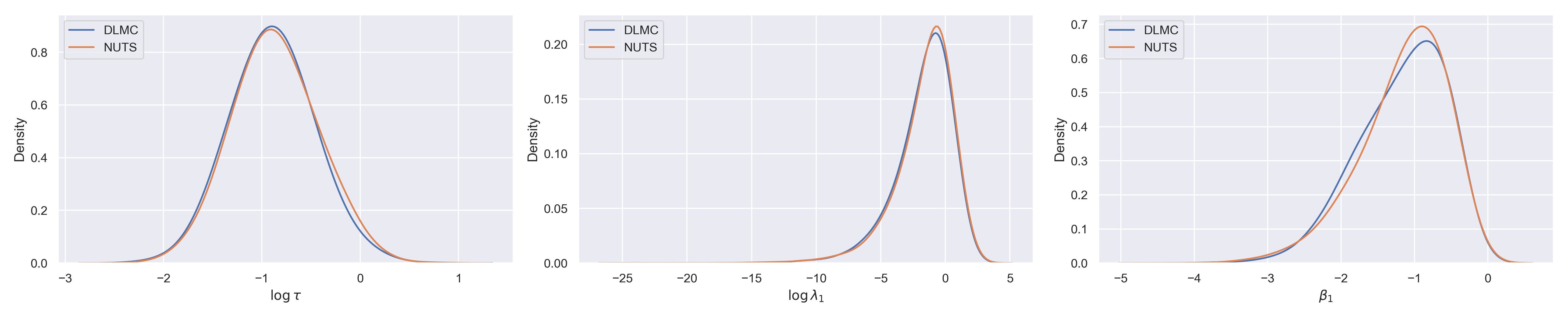}
     \end{subfigure}
     \caption{Results on $d=51$ sparse logistic regression after 100 iterations (blue), compared to the distribution obtained from a very long NUTS run (orange). We show three representative variables, showing good agreement. The corresponding bias-squared is 0.01 (figure 4).}
     \label{fig4}
\end{figure}

\section{Related work}

NFs and transport maps for Bayesian posterior analysis have been explored in \cite{parno2018transport,hoffman2019neutra,DBLP:conf/icml/ArbelMD21}. These approaches use NFs as
latent space preconditioners, using 
standard MCMC methods such as HMC, LMC or MH deployed 
in latent space, where the geometry may 
be more favorable for fast mixing. In \cite{albergo2019flow} NFs are used as an MH transition proposal, giving independent samples. 
DLMC goes beyond these NF applications 
in that it also uses NFs for the density gradient in the deterministic Langevin 
equation. 

DLMC is an ensemble method, in that  
all of the particles inform the NF
density evaluation, which is in 
turn used to update each particle 
position. There have been many other ensemble methods proposed in the 
literature, such as the Affine Invariant Ensemble Sampler \citep{goodman2010ensemble} or Differential Evolution MCMC \citep{ter2006markov}. DLMC 
differs from these methods in that 
it evaluates the particle density 
explicitly, and uses the density and target gradient 
to update the particle evolution. 

Two deterministic algorithms related to DLMC 
that 
have recently been proposed are Stein Variational Gradient Descent (SVGD) \citep{Liu2016stein}, 
which uses direct pair summation
over all particles, and score 
based interacting particles 
\citep{2020Entrp..22..802M}, 
inspired by generative score based models \citep{song2020score}, 
which use the particles to directly estimate the 
density gradient term. DLMC 
differs from these in using NFs 
instead of direct pair 
interactions, which we find requires fewer particles to accurately represent the particle density. 

DLMC is related to  
Variational Inference (e.g., \cite{blei2017variational}) in that both 
approximate the posterior with a 
parametrized normalized function $q(\bi{x})$. However, 
DLMC does not use standard 
variational optimization 
such as KL minimization to optimize
$q(\bi{x})$, 
and does not use $q(\bi{x})$
for the posterior, which is reported
by the particles instead. 
In our experiments 
we observe that the posterior distribution of DLMC
particles 
agrees better with the exact posteriors than the samples from the
NF density. This is expected, since we use the MH 
step to correct the NF. 
DLMC thus
contains components of both 
sampling and optimization:
sampling as optimization in the space of distributions has been 
pursued in many directions since  \cite{jordan1998variational}).

\section{Examples} \label{sec: examples}
The main DLMC application is for 
expensive likelihoods. 
We can simulate this by synthetically 
creating expensive model evaluations, but will also compare the 
performance in terms of the number of 
likelihood evaluations. 
The implementation of DLMC
requires NF and particle evolution. A discussion of implementation details can be found in Appendix \ref{app: dlmc_algo}. For NF
we use 
SINF, which has very few hyper-parameters \citep{dai2021sliced}, is fast, and iterative. The number of layers $L$
can be chosen 
based on cross-validation, 
where we set aside 20\% of the 
samples, and iterate until validation 
data start to diverge from the training data. However, for the $d=1000$ Gaussian (Section \ref{sec: ill_gauss}) we fix $L=5$. 
Typical SINF training
time is of order seconds on a
CPU. 


The implementation of particle evolution requires us to specify the learning rate for the particle updates. At each iteration, we take Adagrad updates in the $\bi{\nabla}(U(\bi{x}(t))-V(\bi{x}(t)))$ direction \citep{JMLR:v12:duchi11a}. We use learning rates between $0.001$ and $0.1$, with smaller learning rates being more robust for targets with complicated geometries such as funnel distributions.

In our 
plots we
focus on the quality of marginal posteriors, 
because that is typically the primary goal 
of Bayesian inference. To quantify the 
quality of the posteriors numerically we follow \cite{hoffman2019neutra} and
show the squared bias of the second moment versus wall clock time or number of gradient evaluations. The
precision of this number is limited by the 
effective number of samples (ESS).
Here we choose 
bias-squared of 0.01 as sufficient in terms of posteriors, and 
corresponds for a Gaussian distribution to an ESS of 200. The number of particles $N$ is a 
hyperparameter that must exceed this 
ESS. Overall we find
there is a tradeoff between number of 
particles and number of iterations, such that 
starting with a larger number of particles
does not always imply a higher overall computational cost, due to the improved NF density estimation resulting in fewer DLMC iterations.

The main baseline we compare against is the No-U-Turn Sampler (NUTS) \citep{hoffman2014no}, an adaptive HMC variant implemented in the NumPyro library \citep{phan2019composable}. It requires
tuning and burn-in, which need to be included
in the computational budget. NUTS is a standard
baseline since it typically 
outperforms other samplers such as LMC, SVGD and SMC. Where we include NUTS as a baseline, we use 500 tuning steps. NUTS has the number of leapfrog steps as a tunable parameter, and when it is one HMC becomes LMC/MALA \citep{girolami2011riemann}.

In some of our experiments we compare default DLMC (denoted serial, S-DLMC) to parallel DLMC (P-DLMC), where 
for the latter we show the wall clock time based on the assumption that we can evaluate 
the likelihood gradient in parallel on $N$ CPU cores. We assume a likelihood gradient cost of 1 minute, and the cost of the NF itself (seconds) is negligible.


\subsection{Gaussian mixture}

Multimodal distributions are a failure mode
for many standard MCMC samplers as they cannot 
move from one peak to another when the potential barrier 
between them is too high.
They need to be augmented with an
annealing procedure, where one slowly morphs the target distribution from the 
prior to the posterior. Annealed Importance Sampling \citep{neal2001annealed} and Sequential MC (SMC) \citep{del2006sequential}
are examples of such procedures. They require  
equilibrating the distribution at every temperature level using standard MCMC, and as a result are significantly slower than standard MCMC. 
DLMC uses a different strategy from annealing, 
but can also handle multi-modal distributions, as long as an NF can 
approximate it. As an example 
we take a Gaussian mixture with 1/3
of the posterior mass in the first peak and 2/3 in 
the second peak, in $d=100$. 

We choose $N=500$, 
drawn from a broad initial prior 
$(-2,2)^d$. The log of the prior to posterior volume ratio is 415, 
which is a challenge for non-gradient based MCMCs, 
as they become very slow in such settings. We apply logit transforms to all the variables such that we sample in an unconstrained space.
DLMC is gradient based and makes rapid 
progress towards the region of high posterior mass. 
The resulting DLMC posterior after 90 iterations for the first two variables
is shown in figure \ref{fig2}, and is in a
near perfect agreement with the true distribution.

\subsection{Rosenbrock function}

The 32-d example comes from a popular variant of the multidimensional 
Rosenbrock function \citep{rosenbrock1960automatic}, 
which is composed of 16 uncorrelated 2-d bananas. The model log-likelihood is
$
\ln L=-\sum_{i=1}^{n/2}\left[\,(g(x_{2i-1})^2-x_{2i})^2/Q+(x_{2i-1}-1)^2\,\right],
Q=0.1, n=32$,
with Gaussian prior $N(0,6^2)$ on all the parameters. We assume $g(x)=x$ and use an initial burn-in phase with $N=10$ for 50 iterations, before upsampling to $N=1000$. 
The posterior for the first two 
variables is shown in figure \ref{fig2}. We 
see that DLMC perfectly matches NUTS. The 
quality of the posterior versus wall clock 
time is shown in figure \ref{fig:funnel_bias2}, 
where we are 
idealizing a data assimilation process by assigning a high
computational cost of 1 minute 
to the evaluation of $g(x)$. 
This allows us to compare serial and 
parallel DLMC. S-DLMC achieves results that 
are competitive with NUTS. P-DLMC can improve
the wall clock time by $N=1000$. 

\subsection{Hierarchical Bayesian analysis}

Hierarchical Bayesian analyses are very common 
in many applications. They often lead to 
funnel type posteriors, which are challenging 
for standard samplers. Moreover, forward
models relating the variance of the latent space
to some parameters of interest can be expensive. 
As an example, 
in cosmology data analysis applications we 
measure Fourier modes and want to 
determine their power spectrum, which in 
turn constrains cosmological parameters. The evaluation of the 
power spectrum as a function of cosmological 
parameters requires an expensive PDE evolution of structure in 
the universe as a function of time \citep{LewisChallinorEtAl00}. Here we choose a simplified 
version where  the power variance $P=g(\theta)$ is 
determined by cosmological parameter $\theta$ and controls the variance of parameters $z_i$. Parameters $z_i$ are not directly observable, 
and instead we observe their 
noisy version with 
a likelihood of the data $\bi{y}$, so a combined prior and likelihood are  
\begin{align}
 p(\theta)= N(0, 3^2),\; 
p(z_i|\theta)= N(0, P=g(\theta)), \; p(y_i|z_i,\theta) = N(z_i, \sigma^2),\; i=1,\ldots,d-1,
\label{funnel_prior}
\end{align}
where $\sigma^2$ is the measurement 
noise and $g(\theta)$ is an expensive function 
that evaluates the power spectrum $P$ as a 
function of $\theta$. To connect 
to Neal's funnel we assume the result 
of the PDE is that the 
power 
spectrum depends exponentially on $\theta$, 
with an approximate relation $P=g(\theta) =\exp(\theta)$, with a 1 minute computational 
cost of performing $g(\theta)$.
The goal is the posterior of parameter $\theta$
marginalized over $z_i$. We use $N=500$. 

In the limit $\sigma^2 \rightarrow \infty$ the setup corresponds to 
Neal's funnel, 
which is deemed problematic for many 
samplers such as HMC \citep{neal2011mcmc}. 
Note however that the DLMC solution in the $\sigma^2 \rightarrow \infty$ is trivial: the data likelihood is non-informative, and 
after 
drawing samples from the prior of equation 
\ref{funnel_prior} the likelihood 
updates of equation \ref{funnel_prior} are zero, so the 
DLMC stopping criterion (stationary posterior) is 
satisfied immediately.

To make the problem closer to 
actual applications in cosmology
we analyze it with a finite 
noise $\sigma^2$, such that 
the 
data are informative of the parameter 
$\theta$. We create a simulated data 
set with the true value 
$\theta_{\rm true}=0$ and use 
$d=101$. The resulting posterior for $\sigma=5$
and $\sigma=0.1$
is compared to NUTS in figure \ref{fig3}. 
We also show the prior (i.e., Neal's funnel) to highlight 
that the posterior is narrow compared
to the prior. 
For $\sigma=5$ we used 15,000 likelihood calls, which 
is orders of 
magnitude fewer likelihood evaluations 
than the NUTS baseline, which struggles to 
converge at all (figure \ref{fig:funnel_bias2}) without the help of a reparametrization or preconditioner \citep{hoffman2019neutra}. 

To highlight the advantage of DLMC with 
parallelization we further show the performance 
of (embarrassingly) parellel DLMC, where we 
evaluate the likelihood update for all 500 particles in parallel, so each 
iteration is $\sim1$ minute. This converges to the required
precision $b^2=10^{-2}$ in 30 minutes. Sequential MCMC methods such as 
NUTS cannot be so efficiently parallelized because they require a hyperparameter tuning and burn-in phase (each of order $10^4$ likelihood calls). This leads to 3-4 orders of magnitude 
larger wall clock time. SMC can take advantage 
of parallelization, but needs a lot of temperature levels and a lot of steps to equilibriate at each temperature level
\citep{del2006sequential,wu2017quantitative}, such that the overall wall clock time is still orders of magnitude larger than parallel DLMC. 

\subsection{Sparse logistic regression}

Hierarchical logistic regression with a 
sparse prior applied to the German credit
dataset is a popular benchmark for sampling 
methods \citep{Dua:2019}. We use the example from \cite{hoffman2019neutra} with $d=51$, applying a log-transform to bounded variables such that we sample in an unconstrained space. This is a challenging example for 
NUTS, with samples requiring on average $\sim 125$ leapfrog steps.

We run an extended burn-in phase with 10 particles to shrink from the prior to posterior volume, using 500 iterations. We then upsample to 5000 samples, with DLMC converging to the correct solution within 10 iterations. Upsampling to lower sample sizes, such as 1000 particles, also converges to the correct solution, but requires more DLMC iterations such that the computational cost is not reduced. The resulting marginals
are shown in figure \ref{fig4} for three representative variables showing good agreement. 

\subsection{Ill conditioned Gaussian}\label{sec: ill_gauss}

To asses the performance of DLMC in high dimensions we consider an ill conditioned Gaussian target in $d=1000$. We construct the target covariance $\Sigma$ by sampling its eigenvalues from $\mathrm{Gamma}(1, 1)$, giving a condition number of $\sim 4000$. The prior and likelihood are then defined as
\begin{equation}
    p(\bi{x})=N(0,\Sigma_\pi=100^2I),\; p(\bi{y}|\bi{x})=N(\bi{x},\Sigma_L=(\Sigma^{-1}-\Sigma_\pi^{-1})^{-1}),
\end{equation}
where the prior and likelihood covariance are chosen such that the posterior covariance is $\Sigma$.

In this high dimensional setting we are limited by the computational cost of the NF, and the ability of the NF to approximate the current particle density. However, we can use a simple flow with 5 layers, to move particles quickly towards the typical set with DLMC. In this experiment we evolved 2000 particles until we satisfied the condition $\mathcal{V}=\sum_{i=1}^d\langle \bi{x}\cdot\bi{\nabla} U(\bi{x})\rangle_{i}\leq 1.1d$, where the sum is over the target dimensions. This virial condition is chosen because initially we have $\mathcal{V}\gg d$, with the particle ensemble satisfying $\mathcal{V}=d$ at equilibrium \citep{29acd3d494044594aea0829ef236aad6}.

For this problem we are unable to achieve low bias with DLMC alone. However, we can treat DLMC as an initialization for HMC i.e., we run DLMC until we reach the virial threshold and then perform parallel HMC with each particle in the NF latent space. DLMC reaches the virial threshold in 28 steps. Running parallel HMC chains in latent space with 10 leapfrog steps and 
the step size targeting an acceptance rate of 0.65, we reach a bias-squared of 0.01 after 5 HMC iterations. This corresponds to $\sim 10^5$ serial likelihood evaluations, and $\sim 100$ parallel evaluations. The resulting particle distributions from DLMC and the subsequent latent space HMC are shown in figure \ref{fig2}. By comparison, NUTS requires $\sim 10^5$ serial likelihood evaluations to reach a bias-squared of 0.01. For this example the serial cost of DLMC+HMC and NUTS is comparable, with DLMC+HMC providing a performance improvement of $\sim 10^3$ when implemented in parallel. In this experiment we have used a very simple implementation of HMC, and the convergence rate could likely be improved by using some ensemble adaptation scheme e.g., \cite{pmlr-v151-hoffman22a}.

\subsection{Ablation studies and comparison to particle deterministic methods}

In Appendix \ref{app: kde_svgd} we 
compare to kernel density estimation based DLMC and SVGD, both of which give inferior results, confirming that the NF is the key ingredient for the success of deterministic methods.

In Appendix \ref{app: ablations} we present results of ablation 
studies. We show DLMC without MH, DLMC without 
NF preconditioning, and pure NF MH without 
DLMC. In all cases the results are inferior to 
DLMC with preconditioning and with MH adjustment. We also show NF with MH without DL is inferior to DLMC. 

\section{Conclusions}

We present a new general purpose Bayesian inference algorithm we call Deterministic Langevin Monte Carlo with Normalizing Flows. 
The method uses the deterministic Langevin 
equation, which requires the density gradient, which 
we propose to be evaluated
using a Normalizing Flow
determined by the positions of all the particles
at a given time. Particle positions
are then updated in the direction of log target density minus log current density gradient. 
To correct for imperfect NF density estimation, we add an MH step. 
The process is repeated until the particle distribution becomes stationary, which happens when the particle density equals the target density. 

In our experiments we find that DLMC significantly outperforms baselines such as LMC, HMC and SVGD, and even more in terms of wall-clock time when likelihoods can be evaluated in parallel on a multi-core computer. 
A possible limitation is that NF training may not learn the 
true target distribution for small $N$. While DLMC with MH does offer some robustness against imperfect density estimation, it can fail in situations 
where the MH acceptance rate is very low, such that it 
cannot reach the stationary distribution in a 
finite number of steps.  This is particularly apparent in $\mathcal{O}(10^3)$ dimensions, where we are limited by the NF training cost and the fidelity of its gradient approximation. Here, convergence on the correct target can be achieved by running DLMC until the particle ensemble reaches the typical set, at which point we perform a short MCMC run in latent space. Indeed, treating DLMC as an initialization for a latent space MCMC in this way can provide some additional robustness through the associated asymptotic guarantees, whilst avoiding the expensive tuning and burn-in process typically required for MCMC.
It would nonetheless be interesting to explore what NF architectures, regularization, and training procedures safeguard against low MH acceptance rates. 

\section*{Acknowledgements}
This material is based upon work supported  by the U.S. Department of Energy, Office of Science, Office of Advanced Scientific Computing Research under Contract No. DE-AC02-05CH11231 at Lawrence Berkeley National Laboratory to enable research for Data-intensive Machine Learning and Analysis. We thank Minas Karamanis, David Nabergoj and James Sullivan for useful discussions.


\bibliography{paper}
\bibliographystyle{plainnat}

\appendix

\section{Checklist}

\begin{enumerate}

\item For all authors...
\begin{enumerate}
  \item Do the main claims made in the abstract and introduction accurately reflect the paper's contributions and scope?
    \answerYes{}
  \item Did you describe the limitations of your work?
    \answerYes{}
  \item Did you discuss any potential negative societal impacts of your work?
    \answerNA{}
  \item Have you read the ethics review guidelines and ensured that your paper conforms to them?
    \answerYes{}
\end{enumerate}

\item If you are including theoretical results...
\begin{enumerate}
  \item Did you state the full set of assumptions of all theoretical results?
    \answerNA{}
        \item Did you include complete proofs of all theoretical results?
    \answerNA{}
\end{enumerate}

\item If you ran experiments...
\begin{enumerate}
  \item Did you include the code, data, and instructions needed to reproduce the main experimental results (either in the supplemental material or as a URL)?
    \answerYes{}
  \item Did you specify all the training details (e.g., data splits, hyperparameters, how they were chosen)?
    \answerYes{}
        \item Did you report error bars (e.g., with respect to the random seed after running experiments multiple times)?
    \answerYes{}
        \item Did you include the total amount of compute and the type of resources used (e.g., type of GPUs, internal cluster, or cloud provider)?
    \answerYes{}
\end{enumerate}

\item If you are using existing assets (e.g., code, data, models) or curating/releasing new assets...
\begin{enumerate}
  \item If your work uses existing assets, did you cite the creators?
    \answerYes{}
  \item Did you mention the license of the assets?
    \answerNo{German credit data taken from public UCI repository.}
  \item Did you include any new assets either in the supplemental material or as a URL?
    \answerYes{Code included as supplemental material.}
  \item Did you discuss whether and how consent was obtained from people whose data you're using/curating?
    \answerNA{}
  \item Did you discuss whether the data you are using/curating contains personally identifiable information or offensive content?
    \answerNA{}
\end{enumerate}

\item If you used crowdsourcing or conducted research with human subjects...
\begin{enumerate}
  \item Did you include the full text of instructions given to participants and screenshots, if applicable?
    \answerNA{}
  \item Did you describe any potential participant risks, with links to Institutional Review Board (IRB) approvals, if applicable?
    \answerNA{}
  \item Did you include the estimated hourly wage paid to participants and the total amount spent on participant compensation?
    \answerNA{}
\end{enumerate}

\end{enumerate}

\section{Appendix: DLMC algorithm implementation}\label{app: dlmc_algo}

The pseudocode for DLMC is given in Algorithm \ref{alg1}. For our implementation of DLMC we initialize all problems with $N_0$ samples from the prior $p(\bi{x})$. This choice is common in implementations of Sequential Monte Carlo, with one of the primary advantages being the corresponding coverage guarantees. The posterior is given by the product of the prior and likelihood, and as such is covered by the prior. This is particularly important when dealing with multimodal targets, where an initialization with poor coverage can lead to missing posterior peaks. This also naturally leads to the first DLMC update (lines 3 to 5 of Algorithm \ref{alg1}), where we move particles along the likelihood gradient direction. At the first step the current particle density is simply the prior, which cancels with the prior term in $U(\bi{x})$. 

For unimodal targets, one can initialize with a very small number of particles (e.g., $\mathcal{O}(10)$), and run some initial burn-in iterations of the primary DLMC loop discussed below. The goal here is to move particles quickly towards the typical set, before drawing new samples from the NF approximation at the end of burn-in to give $N$ particles, with which we continue the primary DLMC loop iterations. In our experience, this is useful in reducing likelihood evaluations when running DLMC in serial mode. However, if users have the computational resource to parallelize their likelihood evaluations, DLMC typically requires a smaller number of total iterations to converge (and hence parallel likelihood evaluations) using a larger ensemble size throughout. In our examples, we used this burn-in strategy for the Rosenbrock function and German credit targets. 

The primary DLMC loop is given in lines 6 to 19 of Algorithm \ref{alg1}. At the beginning of each DLMC iteration we retrain the NF on the current particle positions, giving our estimate for $q(\bi{x}(t))=\exp(-V(\bi{x}(t)))$, along with the corresponding latent space map $\bi{z}=f(\bi{x}(t))$.  The DL updates can then either be performed in the NF latent space, or in the original data space. In both cases, we have some gradient term $\bi{g}(t)$ which we use to perform an optimization update on our particle ensemble. As discussed in Section \ref{sec: langevin_fp}, any optimization algorithm can be used to perform this update. For our implementation we use Adagrad updates at each DLMC iteration. In principle, one can perform multiple optimization updates before updating the NF. However, for our purposes we use a single Adagrad update at each DLMC iteration before updating the NF. 

The latent space update can be viewed as a preconditioned update. A typical approach to preconditioning (e.g., through mass matrix adaptation in HMC) is to obtain some estimate of the target covariance. In practical implementations this is often taken as a diagonal preconditioner in order to avoid the inversion of a dense covariance estimate in high dimensions. In this case, the preconditioner applies a global scaling (and rotation in the case of a dense preconditioner), which helps to accelerate convergence but cannot account for local variations in spatial curvature. As an alternative to these global preconditioners, one can employ Riemann Manifold Monte Carlo (RMMC), where a position dependent metric accounts for the local curvature of the target \cite{girolami2011riemann}. However, the requirement to evaluate this position dependent metric, and the corresponding implicit integrator for the RMMC updates, renders the stable numerical implementation of the method challenging.

For DLMC, we use the NF fit to map the particle ensemble to the NF latent space. At each iteration, the NF seeks to map the current particle density to the standard Gaussian. For complex target geometries, this has the effect of mapping spatially varying curvature to a latent space with constant and isotropic curvature. The latent space target density in equation \ref{pz} is given by the change of variables formula, which contains a Jacobian term that is easy to evaluate for the NF by construction. We find that performing DLMC updates in latent space significantly reduces the number of iterations required for convergence.

The final step in the primary DLMC loop is the independent MH update. Whilst this is not strictly necessary for DLMC, we find the addition is important for correcting any imperfect NF density estimation and eliminating particles that become stuck far from the typical set. The impact of dropping the MH update is illustrated in figure \ref{fig:funnel_noMH}, where we run DLMC on the hierarchical funnel without MH updates. In this case, whilst most particles move towards the target, some become stuck far from the typical set preventing proper convergence.

\section{Appendix: DLMC with kernel density estimation and SVGD}\label{app: kde_svgd}

To solve for the deterministic 
Langevin equation we need $\bi{\nabla}\ln q(\bi{x}(t))$.
Instead of NFs we can use kernel 
density estimation (KDE) for the 
density gradient (see also \cite{2020Entrp..22..802M}). We can define a
general normalized kernel density $k(\bi{x},\bi{x}')$. A special case is a 
Gaussian (a radial basis function),
\begin{equation}
k_G(\bi{x},\bi{x}')=(2\pi \sigma^2)^{-1/2}\exp\left(-\frac{|\bi{x}-\bi{x}'|^2}{2\sigma^2}\right),   
\end{equation}
where $\sigma$ is the kernel 
width.  
The KDE density at $\bi{x}$ is 
\begin{equation}
q(\bi{x})=\frac{1}{N} \sum_{j=1}^N k(\bi{x},\bi{x}^j).   
\end{equation}
Its logarithmic gradient at 
particle $\bi{x}^i$ is
\begin{equation}
\nabla V(\bi{x}^i)=\frac{ -\sum_{j=1}^N \nabla_{\bi{x}^i}k(\bi{x}^i,\bi{x}^j)}{ \sum_{j=1}^N k(\bi{x}^i,\bi{x}^j)}. 
\label{gradvkde}
\end{equation}

For a Gaussian kernel this becomes
\begin{equation}
\nabla V(\bi{x}^i)=\frac{ \sum_{j} k_G(\bi{x}^i,\bi{x}^j)[\bi{x}^j-\bi{x}^i]}{ \sigma^2\sum_{j} k_G(\bi{x}^i,\bi{x}^j)}.  
\end{equation}
Note that the self-interaction 
term $i=j$ vanishes. 
We will call this version particle-particle DLMC (PP-DLMC), 
since it is based on direct pair 
summation over all particles. 
The gradient points towards the nearest 
particles, weighted by the 
kernel $k$. The hyperparameter $\sigma^2$ controls the number of 
particles that will contribute to the density gradient: for $\sigma^2 \rightarrow 0$ this is completely 
determined by the nearest particle, 
and the gradient explodes, 
while for $\sigma^2 \rightarrow \infty$
this is determined equally by all the particles, and the gradient goes to 0. 
 
\begin{figure*}[t]
     \centering
       \begin{subfigure}[t]{0.45\linewidth}
         \includegraphics[width=\textwidth]{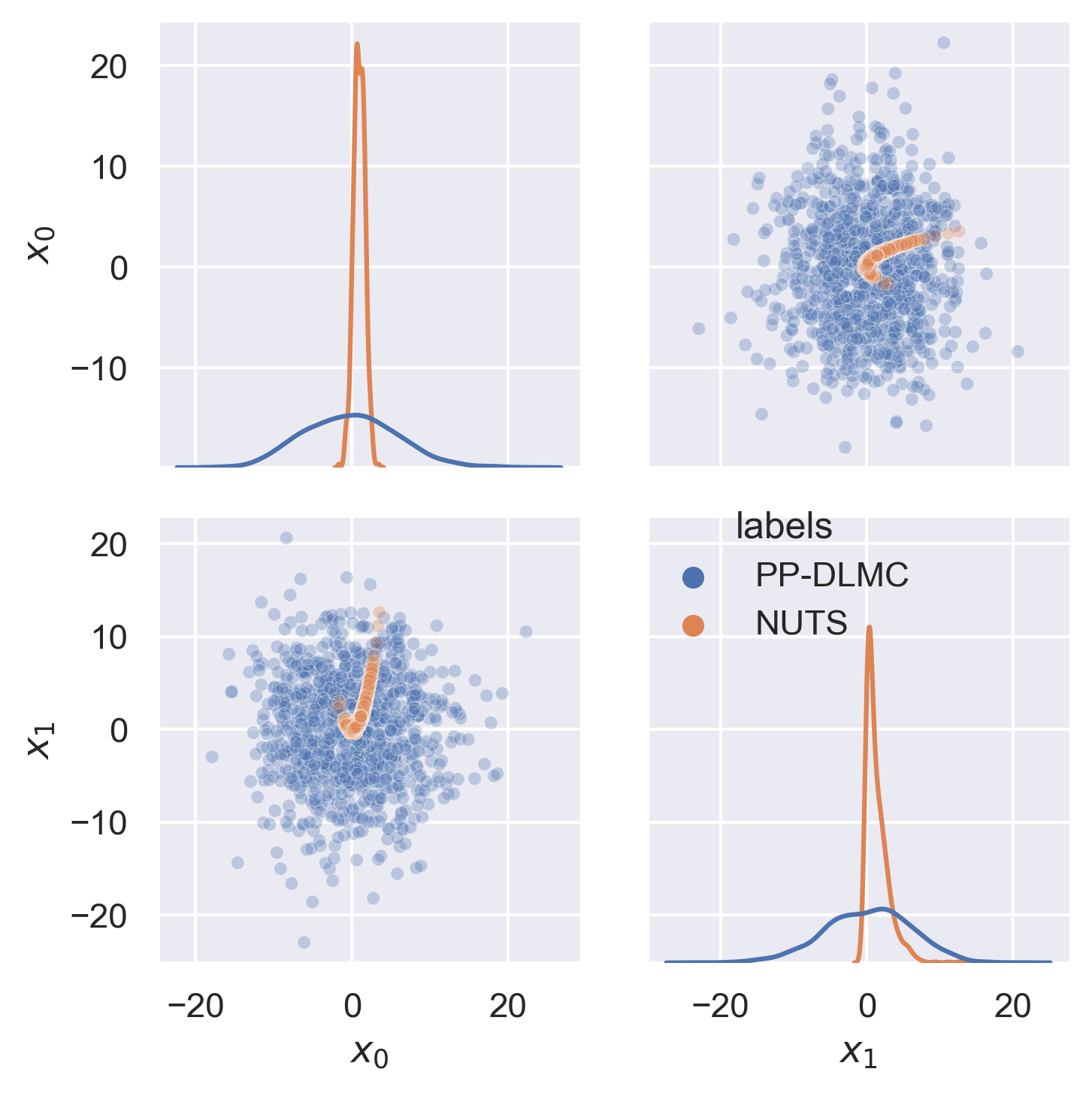}
         \caption{Particle-particle DLMC}
     \end{subfigure}
     \caption{Marginal distributions for a $d=32$ Rosenbrock for PP-DLMC, DLMC with a particle-based kernel density estimate. This can be compared to figure \ref{fig2}. We see that PP-DLMC does 
     not converge to the true distribution due to the high dimensionality and narrow posterior shape.}
     \label{fig:pp}
      \vskip -0.1in
\end{figure*}

 Results for PP-DLMC with optimized $\sigma$ are shown in 
 figure \ref{fig:pp} for the Rosenbrock
 distribution. We were unable to obtain 
 settings that converged to the correct 
 answer. We interpret these results 
 as evidence that the KDE approximation 
 is inferior to the NF approximation in high dimensions 
 and/or for challenging distributions, 
 which is known in other contexts 
 (see e.g., \cite{dai2021sliced} for 
 comparison between KDE and NFs on standard 
 UCI datasets).

The closest method to PP-DLMC is 
Stein Variational Gradient Descent (SVGD) \citep{Liu2016stein}, which 
evaluates the updates as 
\begin{align}
    \frac{N\Delta \bi{x}^i}{\Delta t}=\sum_{j=1}^N\left[-\bi{\nabla}_{\bi{x}^j} U(\bi{x}^j)k(\bi{x}^i,\bi{x}^j)+\nabla_{\bi{x}^j}k(\bi{x}^i,\bi{x}^j)\right]. 
    \label{Lan9}
\end{align}
We see that SVGD does not use the 
target gradient at the particle position $\bi{x}^i$,
but instead averages the target gradients 
over all the other particles, 
weighted by the kernel. While this 
leads to a different dynamics from PP-DLMC, 
we expect the two to be qualitatively 
similar. 

\begin{figure*}[t]
     \centering
     \begin{subfigure}[t]{0.3\linewidth}
         \includegraphics[width=\textwidth]{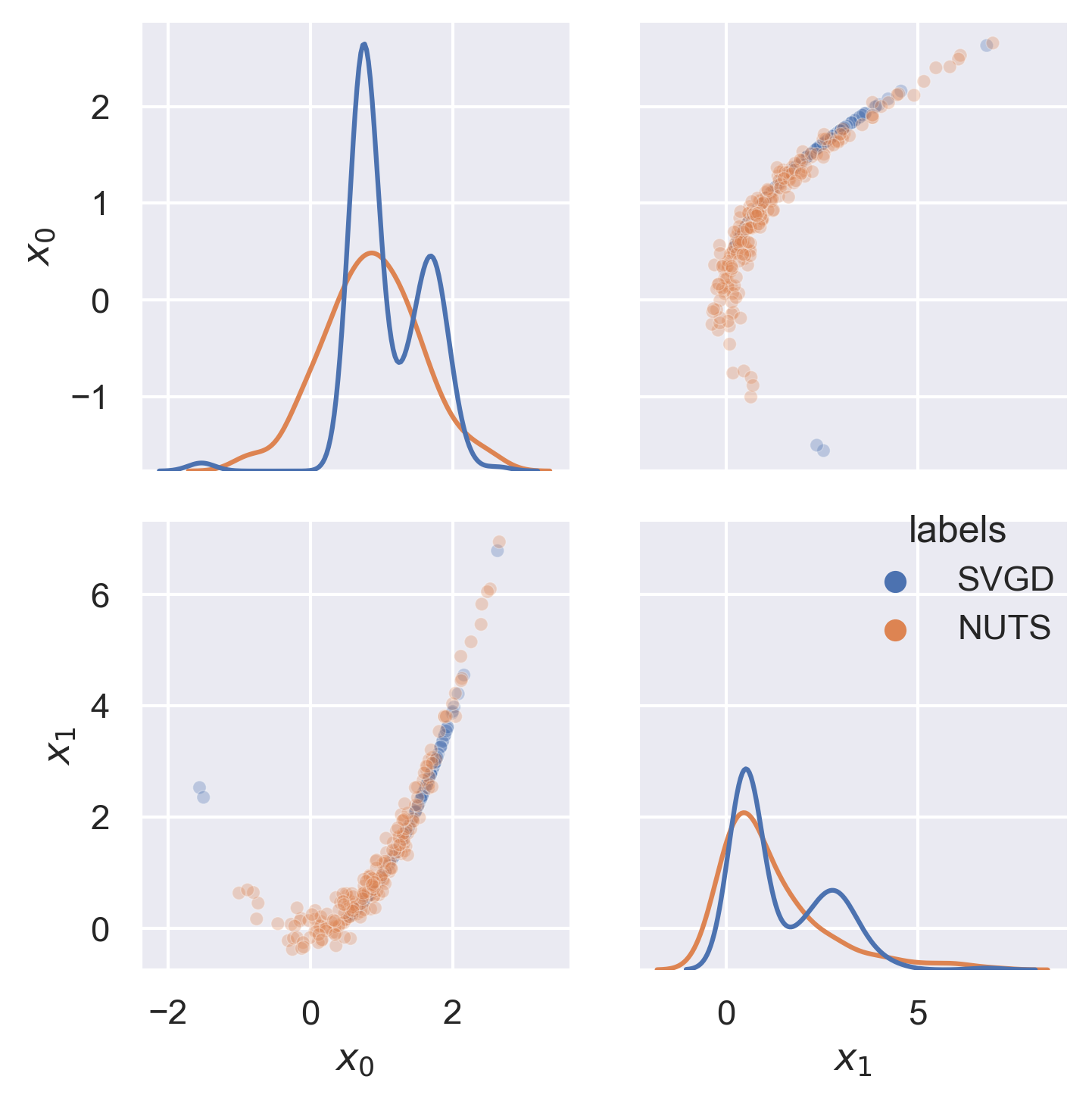}
         \caption{Rosenbrock}
     \end{subfigure}
      \begin{subfigure}[t]{0.3\linewidth}
         \includegraphics[width=\textwidth]{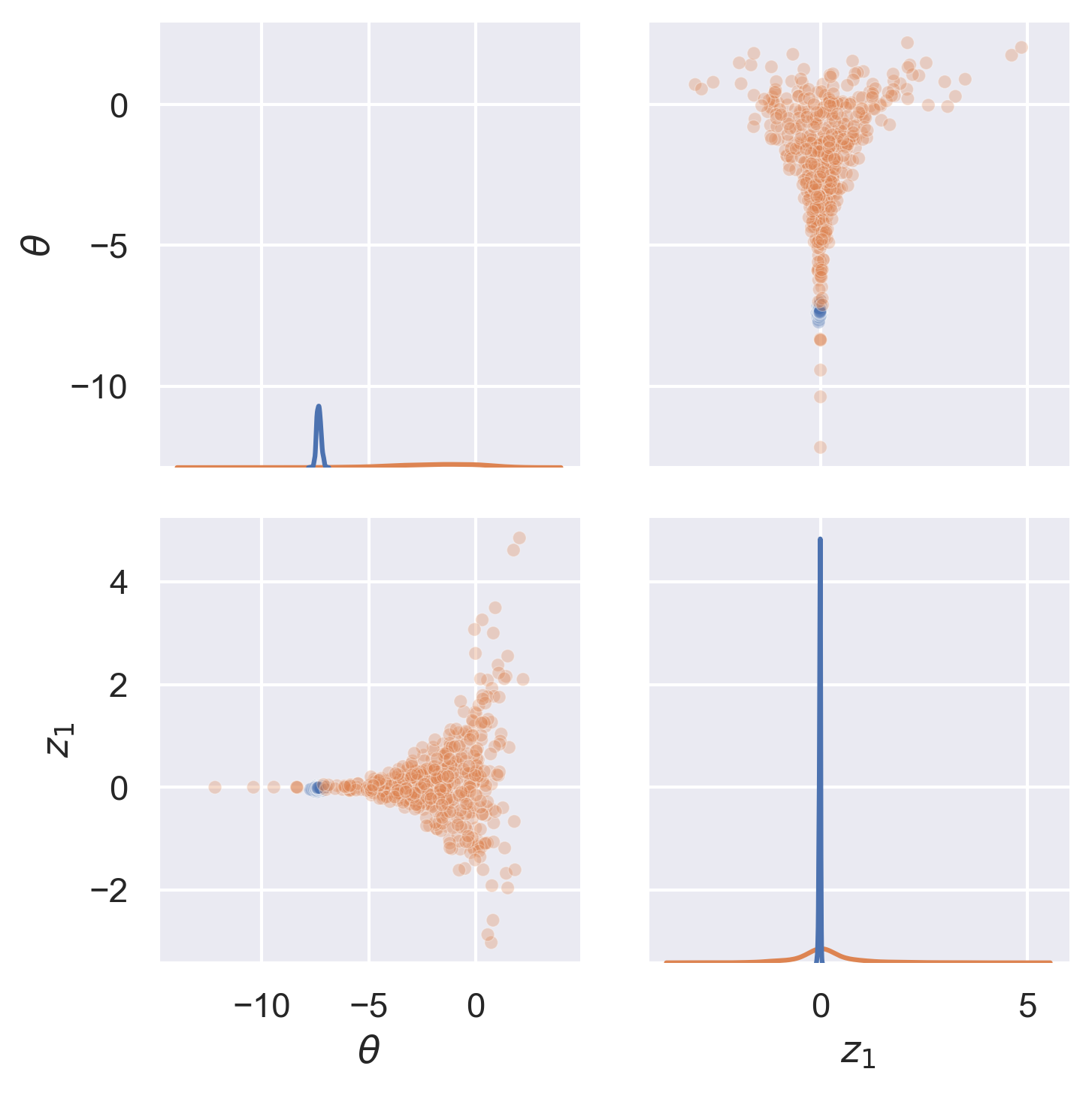}
         \caption{funnel $\sigma=5$}
     \end{subfigure}
       \begin{subfigure}[t]{0.3\linewidth}
         \includegraphics[width=\textwidth]{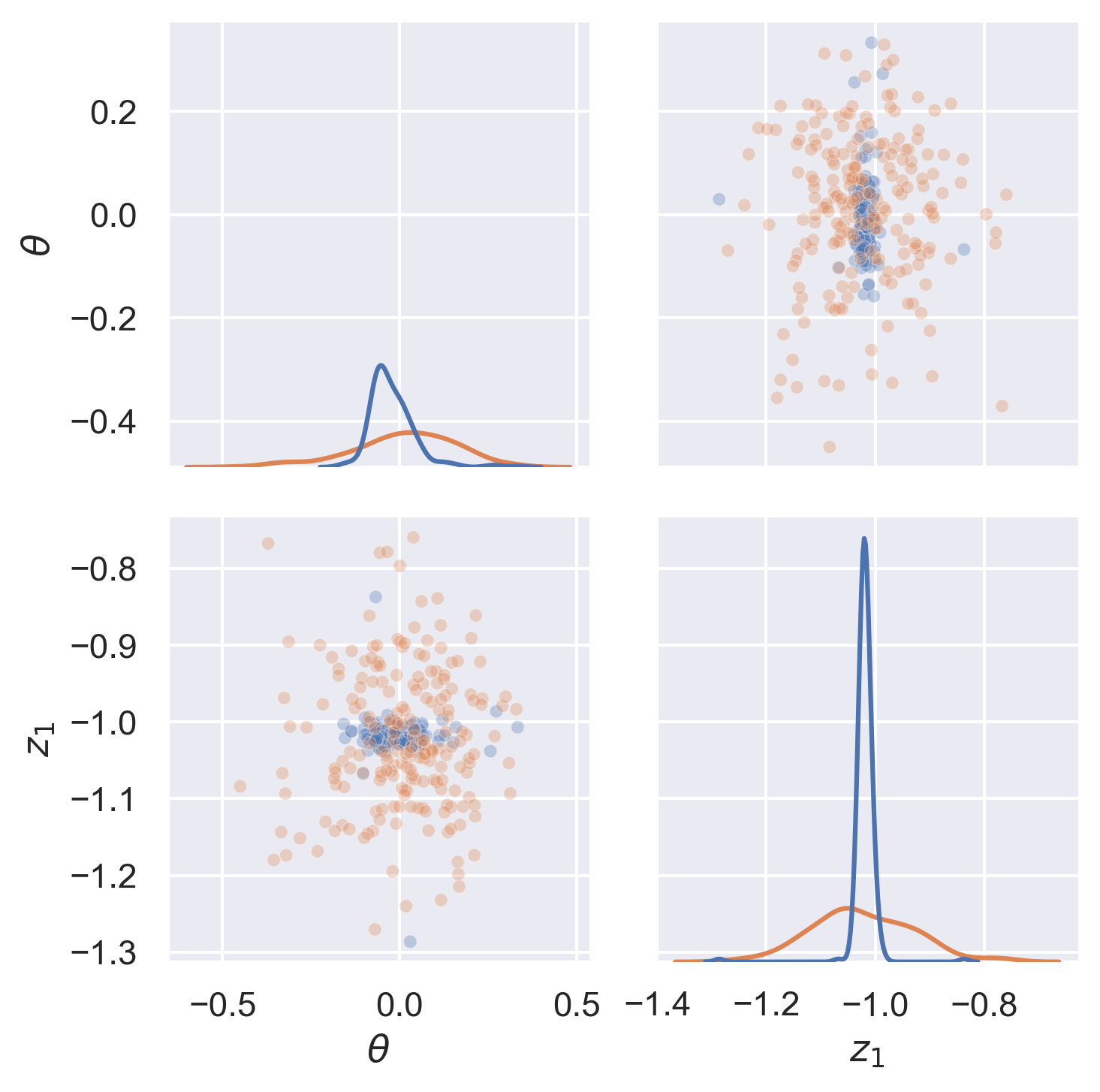}
         \caption{funnel $\sigma=0.1$}
     \end{subfigure}
      \begin{subfigure}[t]{0.9\linewidth}
         \includegraphics[width=\textwidth]{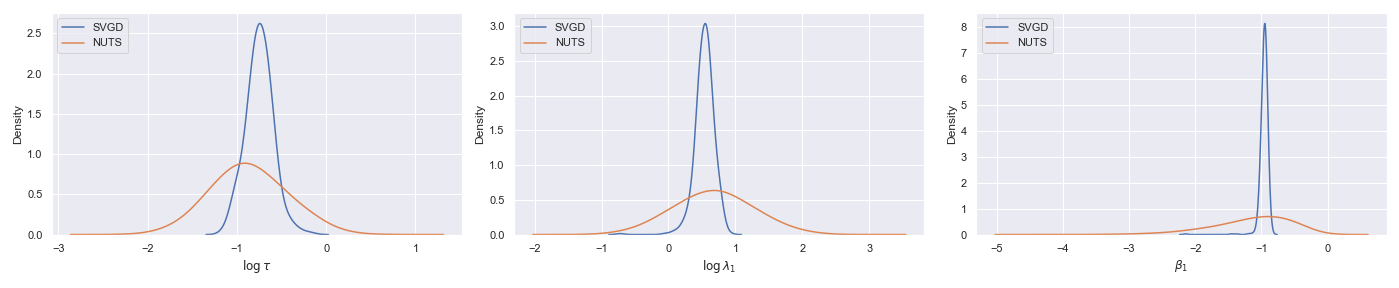}
         \caption{sparse logistic regression}
     \end{subfigure}
     \caption{Results at iteration 10000 for SVGD on the Rosenbrock (1000 particles), $\sigma=5$ funnel (500 particles), $\sigma=0.1$ funnel (200 particles), and sparse logisitic regression (500 particles) models. In all cases we are unable to converge on the correct solution.}
     \label{fig:svgd}
      \vskip -0.1in
\end{figure*}

We test SVGD against the Rosenbrock, funnel and sparse logistic regression models.  We show example results from SVGD in figure \ref{fig:svgd}.  In all cases we are unable to converge on the correct solution. We interpret 
these results as evidence that SVGD performance significantly degrades in 
high dimensional and/or challenging examples such as these, 
consistent with the argument above that 
SVGD is closely related to PP-DLMC, which is based on KDE, and KDE is a 
poor density estimator in high 
dimensions or for challenging distributions, 
relative to NFs. 

\section{Appendix: ablation tests}\label{app: ablations}

\begin{figure*}[t]
     \centering
      \begin{subfigure}[t]{0.45\linewidth}
         \includegraphics[width=\textwidth]{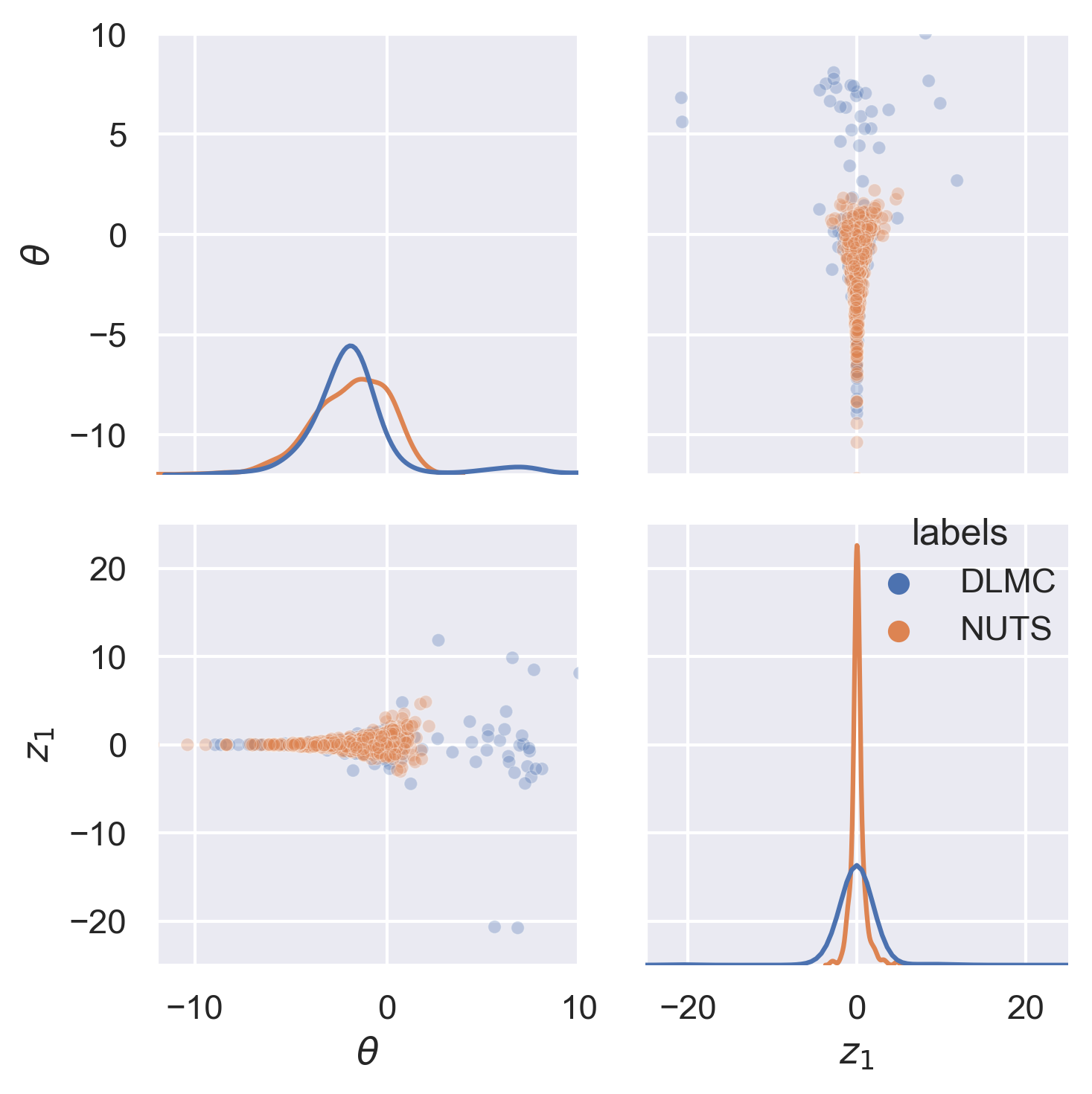}
     \end{subfigure}
     \caption{DLMC sample distribution for the $d=101$ funnel with $\sigma=5$, obtained without applying MH corrections. The MH step is needed for proper convergence.}
     \label{fig:funnel_noMH}
      \vskip -0.1in
\end{figure*}

In figure \ref{fig:funnel_noMH} we 
evaluate DLMC without MH for the hierarchical Bayesian example of figure \ref{fig3}, with 
$\sigma=5$, at the same number of gradient 
evaluations. We see that without MH the distribution of 
$z_1$ and $\theta$ is broader than the true 
distribution. As suggested
in the main text, MH is helpful to 
remove the worst performers and replace 
them with particles at higher density. 

\begin{figure}[t]
     \centering
     \hskip 0.3in
      \begin{subfigure}[t]{\linewidth}
         \includegraphics[width=0.9\textwidth]{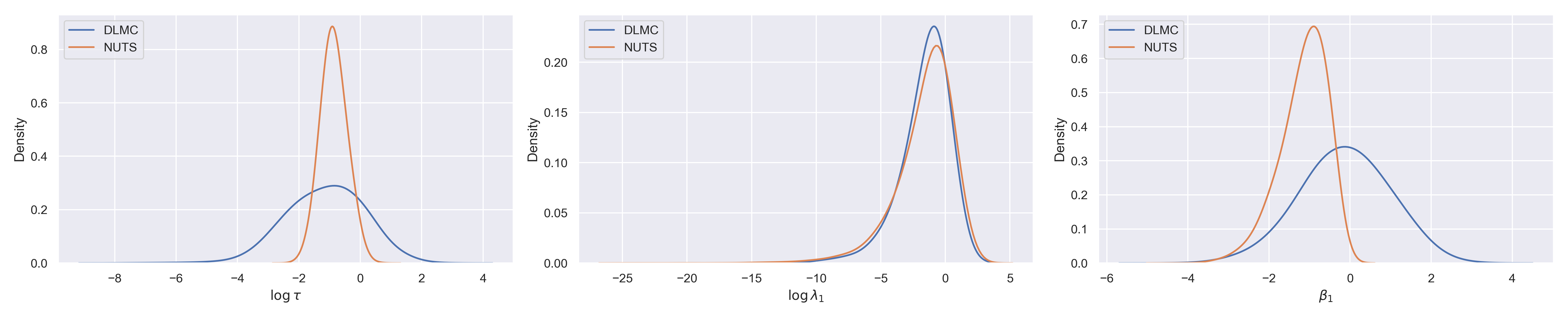}
     \end{subfigure}
     \caption{Results on $d=51$ sparse logistic regression, without NF preconditioning after the same number of evaluations as figure \ref{fig4}. We observe slower convergence 
     without NF preconditioning.}
     \label{fig4_noz}
\end{figure}

In figure \ref{fig4_noz} we show 
results on $d=51$ sparse logistic regression, without NF preconditioning after the same number of evaluations as figure \ref{fig4}. We observe slower convergence 
     without NF preconditioning.

\begin{figure*}[t]
     \centering
      \begin{subfigure}[t]{0.45\linewidth}
         \includegraphics[width=\textwidth]{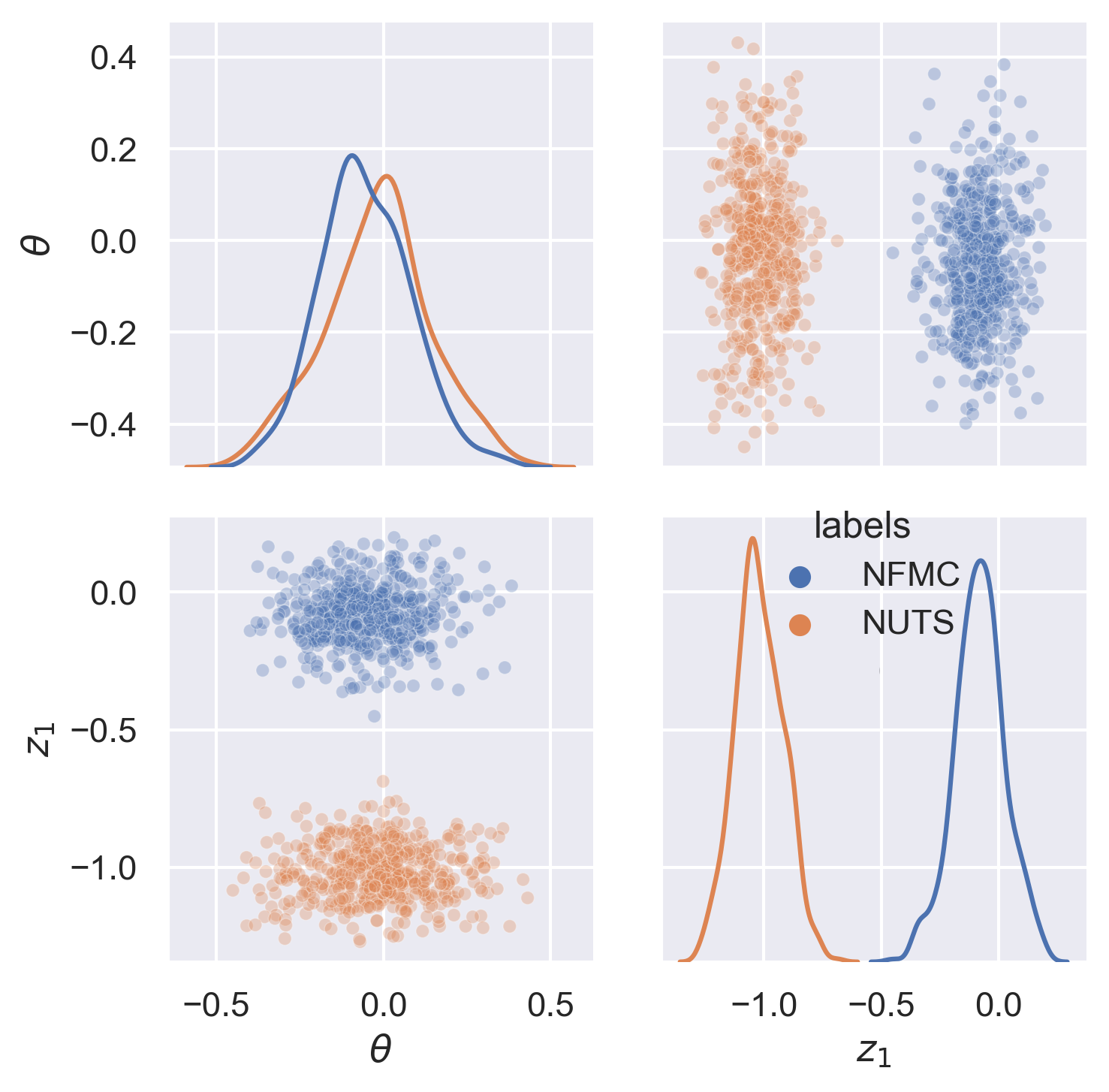}
     \end{subfigure}
     \caption{DLMC sample distribution for the $d=101$ funnel with $\sigma=0.1$, obtained by iteratively fitting an NF to the sample density and applying MH updates using the NF as a proposal distribution for 200 iterations, with 500 particles. We observe 
     very slow convergence in the absence of the DL step.}
     \label{fig:noDL}
      \vskip -0.1in
\end{figure*}

In figure \ref{fig:noDL} we show the DLMC sample distribution for the $d=101$ funnel with $\sigma=0.1$, obtained by iteratively fitting an NF to the sample density and applying MH updates only, using the NF as a proposal distribution for 200 iterations, with 500 particles. We observe very slow convergence in the absence of a DL step.

\end{document}